\documentclass{article} % For LaTeX2e
\usepackage{iclr2024_conference,times}

\usepackage{amsmath,amsfonts,bm}

\def\eqref#1{equation~\ref{#1}}
\def\ri{{\textnormal{i}}}

\DeclareMathAlphabet{\mathsfit}{\encodingdefault}{\sfdefault}{m}{sl}
\SetMathAlphabet{\mathsfit}{bold}{\encodingdefault}{\sfdefault}{bx}{n}

\DeclareMathOperator*{\argmax}{arg\,max}
\DeclareMathOperator*{\argmin}{arg\,min}

\usepackage{hyperref}
\usepackage{url}
\usepackage{graphicx}
\usepackage{subfigure}
\usepackage{multirow}
\usepackage{multicol}
\usepackage{bbm}
\usepackage{float}
\usepackage{tablefootnote}

\usepackage{amsmath}
\usepackage{amssymb}
\usepackage{mathtools}
\usepackage{amsthm}
\usepackage{caption}
\usepackage{algorithmic}
\usepackage{algorithm}
\usepackage[capitalize,noabbrev]{cleveref}
\usepackage{comment}
\usepackage{thmtools}
\usepackage{thm-restate}
\usepackage{graphicx}
\usepackage{wrapfig}
\theoremstyle{plain}
\newtheorem{theorem}{Theorem}[section]

\newtheorem{lemma}[theorem]{Lemma}

\theoremstyle{definition}

\theoremstyle{remark}

\RequirePackage{algorithm}
\RequirePackage{algorithmic}
\usepackage[textsize=tiny]{todonotes}

\title{Leveraging Uncertainty Estimates to Improve Classifier Performance}

\author{Gundeep Arora, Srujana Merugu, Anoop Saladi, Rajeev Rastogi \\
Amazon\\
\texttt{gundeepa@amazon.com} \\
}

\begin{document}

\maketitle

%% ========================================================
% \newcommand\R{\mathbb{R}}
\newcommand\U{\mathbb{U}}
\newcommand\eS{\mathbb{S}} % why did we do this
\newcommand\I{\mathbb{I}}
\newcommand\1{\mathbbm{1}}

%% ========================================================
%% Some useful commands. Note the dangerous \r command!
%% ========================================================
\newcommand{\bd}{\boldsymbol}
\newcommand{\mf}{\mathbf}

\renewcommand{\a}{\mathbf{a}}
\renewcommand{\c}{\mathbf{c}}
\renewcommand{\r}{\mathbf{r}} % since \r was already defined in altex
\renewcommand{\u}{\mathbf{u}}
\renewcommand{\v}{\mathbf{v}}
\newcommand{\s}{\mathbf{s}}
\newcommand{\w}{\mathbf{w}}
\newcommand{\x}{\mathbf{x}}
\newcommand{\y}{\mathbf{y}}
\newcommand{\z}{\mathbf{z}} 
\newcommand{\p}{\mathbf{p}}
\newcommand{\myb}{\mathbf{b}}
\newcommand{\q}{\mathbf{q}}

% \algnewcommand\algorithmicreturn{\textbf{return}}
% \algnewcommand\RETURN{\item[\algorithmicreturn]}

%----------------------------------------
%      THE SETS
%----------------------------------------
\newcommand{\cC}{{\cal C}}
\newcommand{\cD}{{\cal D}}
\newcommand{\cE}{{\cal E}}
\newcommand{\cU}{{\cal U}}
\newcommand{\cL}{{\cal L}}
\newcommand{\cM}{{\cal M}}
\newcommand{\cP}{{\cal P}}
\newcommand{\cQ}{{\cal Q}}
\newcommand{\cS}{{\cal S}}
\newcommand{\cT}{{\cal T}}
\newcommand{\cX}{{\cal X}}
\newcommand{\cY}{{\cal Y}}
\newcommand{\cZ}{{\cal Z}}
\newcommand{\cF}{{\cal F}}
\newcommand{\cB}{{\cal B}}
\newcommand{\cA}{{\cal A}}

%--------------------------------------------
%         THE VARIABLES & APPROXIMATES
%--------------------------------------------
\newcommand{\bM}{\mathbf{M}}
\newcommand{\bW}{\mathbf{W}}
\newcommand{\bA}{\mathbf{A}}
\newcommand{\bB}{\mathbf{B}}
\newcommand{\bR}{\mathbf{R}}
\newcommand{\bC}{\mathbf{C}}
\newcommand{\bE}{\mathbf{E}}
\newcommand{\bZ}{\mathbf{Z}}
\newcommand{\bG}{\mathbf{G}}

\newcommand{\hZ}{\hat{Z}}
\newcommand{\hM}{\hat{M}}
\newcommand{\hA}{\hat{A}}
\newcommand{\hC}{\hat{C}}
\newcommand{\hU}{\hat{U}}
\newcommand{\hV}{\hat{V}}
\newcommand{\hX}{\hat{X}}
\newcommand{\hY}{\hat{Y}}
\newcommand{\hG}{\hat{G}}

\newcommand{\hz}{\hat{z}}
\newcommand{\hu}{\hat{u}}
\newcommand{\hv}{\hat{v}}
\newcommand{\hx}{\hat{x}}
\newcommand{\hy}{\hat{y}}

%---------MISC-----------------------------
\newcommand{\pr}{\mathbf{P}}
\newcommand{\xx}{\boldsymbol{\chi}}
\newcommand{\Th}{\boldsymbol{\theta}}
\newcommand{\Eta}{\boldsymbol{\eta}}

\newcommand{\norm}[2]{\ensuremath \|#1\|_{#2}}
\newcommand{\myref}[1]{(\ref{#1})}
\newcommand{\del}[2]{\frac{\partial #1}{\partial #2}}

\newcommand{\proofsketch}{\noindent{\itshape Proof Sketch:}\hspace*{1em}}

\begin{abstract}
%O: classification important
%O: typical approach
%O: problems - estimation bias and dependence on uncertainty
%O: summary of contributions

%SM: requires shortening!
Binary classification involves predicting the label of an instance based on whether the model score for the positive class exceeds a threshold chosen based on the application requirements (e.g., maximizing recall for a precision bound). However, model scores are often not aligned with the true positivity rate. This is especially true when the training involves a differential sampling across classes or there is distributional drift between train and test settings. In this paper, we provide theoretical analysis and empirical evidence of the dependence of model score estimation bias on both uncertainty and score itself. Further, we  formulate the decision boundary selection in terms of both model score and uncertainty, prove that it is NP-hard, and present  algorithms  based on dynamic programming and isotonic regression.  Evaluation of the proposed algorithms on three real-world datasets yield  25\%-40\%  gain in recall at high precision bounds over the traditional approach of using model score alone, highlighting the benefits of leveraging uncertainty.
\end{abstract}

\section{Introduction}
\label{sec:intro}
%O: motivate imbalanced binary classification with examples
%O: max recall with precision bound or opposite since our approach  can handle both
Many real-world applications such as fraud detection and medical diagnosis can be framed as binary classification problems, with the positive class instances corresponding to fraudulent cases and disease diagnoses, respectively.
%that are often rare relative to the negative instances.  
When the predicted labels from the classification models are used to drive strict actions, e.g., blocking fraudulent orders and risky treatments, it is critical to minimize the impact of erroneous predictions. This warrants careful selection of the class decision boundary  using the model output while managing the precision-recall trade-off as per application needs. 
%Often, in such scenarios, the decision boundary selection is  posed in terms of maximizing recall such that precision exceeds a specified bound or vice versa.

%O: typical approach for handling imbalance and the decision boundary construction
Typically, one learns a classification model from a training dataset. The class posterior distribution from the model is then used to obtain the precision-recall (PR) curve  on a hold-out dataset with distribution similar to the deployment setting. Depending on the application need, e.g., maximizing recall subject to a precision bound, a suitable operating point on the PR curve is identified to construct the decision boundary. The calibration  on the hold-out set is especially important for applications  with severe class imbalance since it is a common practice to downsample the majority\footnote{\footnotesize Without loss of generality, we assume that the downsampling is performed on the -ve class (label=$0$) and the model score refers to +ve class (label=$1$) probability.} class during model training. This approach of downsampling followed by calibration on hold-out set is known to  both improve model accuracy and reduce computational effort~\citep{downsampling}. 
%overhead due to smaller data size ~\citep{downsampling}. 

%O: existing limitations 
A key limitation of the above widely used approach is that the decision boundary is constructed solely based on the classification model score and  does not account for the prediction uncertainty, which has been the subject of active research in recent years ~\citep{ZHOU2022449,NEURIPS2018_a981f2b7}. \textit{A natural question that emerges is  whether two regions with similar scores but different uncertainty estimates should be treated identically when constructing the decision boundary.}  
%There have been several studies that
Recent studies point to potential benefits of combining model score with estimates of aleatoric (i.e., intrinsic to the input) and epistemic uncertainty (i.e., due to model or data inadequacy) \citep{bayesiandlcv} for specialized settings~\citep{thresholding2022} or via  heuristic approaches ~\citep{heuristic2d2022}. However, there does not exist  in-depth analysis on why incorporating uncertainty leads to better classification, and how it can be adapted to any generic model in a post-hoc setting. 

\begin{wrapfigure}{l}{0.47\textwidth}
% \hfill
\vspace{-3mm}
\subfigure[]{\includegraphics[width=0.47\linewidth,height=2.8cm]{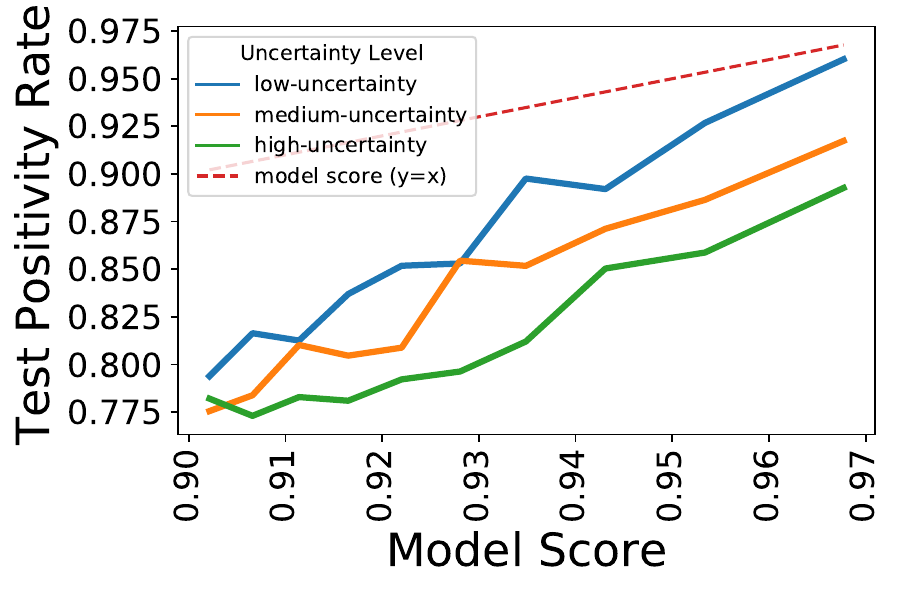}}
% \hfill
\begin{subfigure}[]
{\includegraphics[width=0.49\linewidth,height=2.8cm]{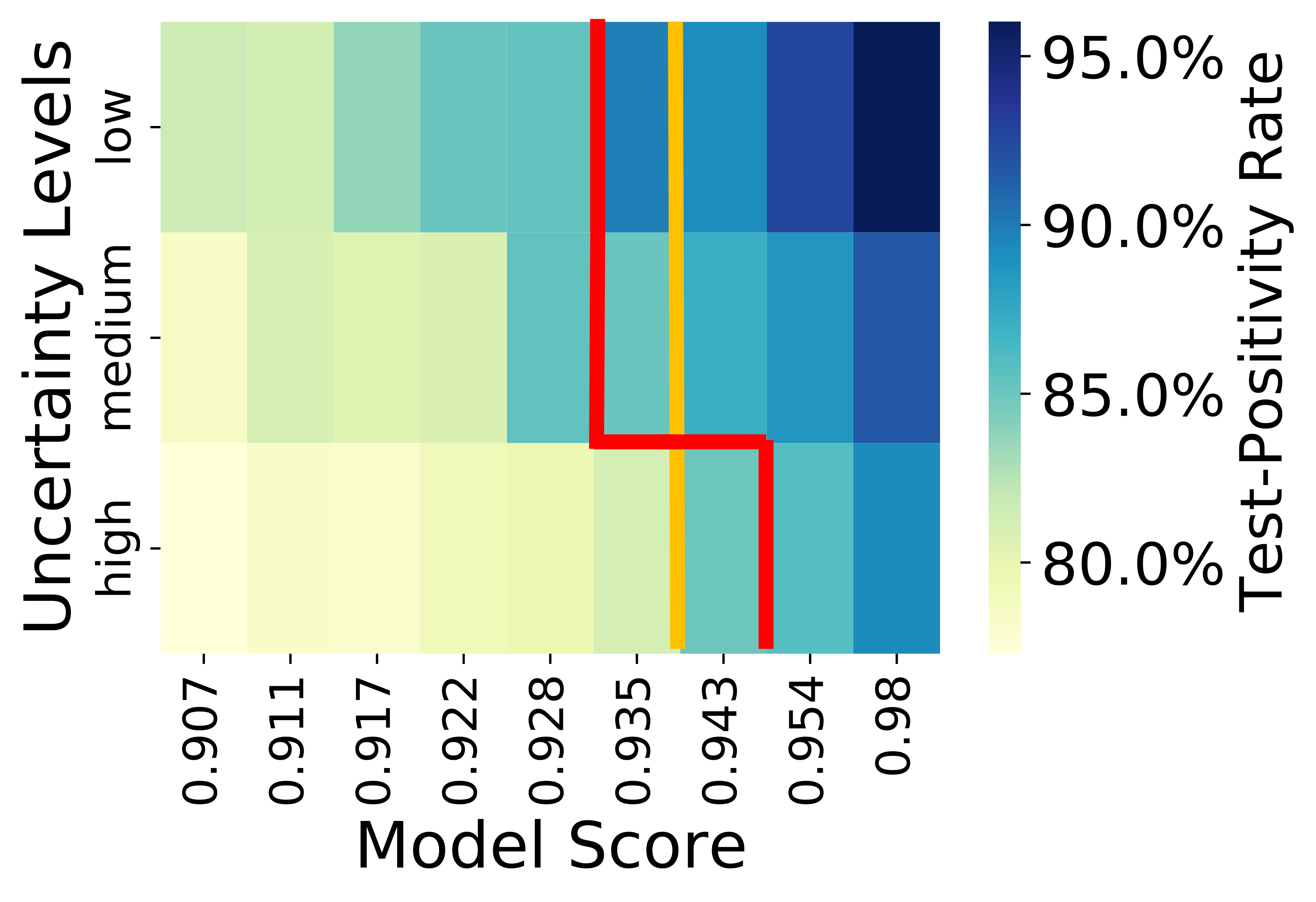}}
\end{subfigure}
%\todo{add decision boundary lines}
\vspace{-3mm}
% \hfill
%\vskip -0.2in
\caption{\footnotesize (a) Test positivity rate vs. model score for different uncertainty levels on \texttt{Criteo} with $33\%$ undersampling of negatives during training. (b) Heatmap of test positivity 
%rate
for different score and uncertainty ranges. Proposed method(red) yields 
%superior 
better 
recall over vanilla score-based threshold (yellow).}
\label{fig:teaser-results}
\vspace{-2mm}
\end{wrapfigure}
% % teaser image
% \begin{figure}[!htbp]
% \vspace{-8mm}
% \subfloat{%
%   \includegraphics[clip,width=0.8\columnwidth,height=5cm]{images/teaser/mnist/lineplot_BinNest_score_unc_test.pdf}%
% }
% \vspace{-8mm}
% \subfloat{%
%   \includegraphics[clip,width=0.8\columnwidth,height=5cm]{images/teaser/mnist/lineplot_BinNest_score_unc_bias_test.pdf}%
% }
% \vspace{-8mm}
% \subfloat{%
%   \includegraphics[clip,width=0.9\columnwidth,height=5cm]{images/teaser/mnist/hm_mnist_teaser_db.png}%
% }
% % \vspace{-5mm}
% \caption{The plots demonstrate the relation between model score's level of over-estimation and the uncertainty range within which the samples lie, because of the differential sampling of negative class in train and test data. The heatmap plots the empirical positivity rate with different score and uncertainty ranges. Our proposed approach (red bar) results in incremental recall as compared to vanilla score based threshold (blue)}
% \label{fig:teaser-results}
% \end{figure}
% \begin{figure}
%     \includegraphics[width=.45\textwidth]{images/teaser/mnist/lineplot_BinNest_score_unc_test.pdf}
%     \\[\smallskipamount]
%     \includegraphics[width=.45\textwidth]{images/teaser/mnist/lineplot_BinNest_score_unc_bias_test.pdf}
%     \\\hspace{-10mm}
%     \includegraphics[width=.45\textwidth]{images/teaser/mnist/heatmap_BinNest_score_unc_test.pdf}
%     \caption{Some images}\label{fig:foobar}
% \end{figure}

%O: research questions 
In this paper, we focus on  binary classification with emphasis on the case where class imbalance  requires differential sampling during training. 
We investigate four questions:\\
 \noindent {\bf RQ1:} Does model score estimation bias (deviation from test positivity)  depend on uncertainty? \\
 %Can this finding help improve decision boundary selection?\\
 \noindent {\bf RQ2:} If so, how can we construct an optimal 2D decision boundary using both model score and uncertainty and what is the relative efficacy?\\
 %of different solutions?\\
 \noindent {\bf RQ3:} Under what settings (e.g., undersampling of negative class, precision range) do we gain the most from incorporating uncertainty? \\
 \noindent {\bf RQ4:} Do uncertainty estimates also aid in better calibration of class probabilities?
%We study the dependence of score estimation bias on uncertainty, and build on the findings to improve decision boundary selection. Particularly, we explore various algorithmic solutions to construct optimal 2D decision boundary using both model score and uncertainty under multiple problem settings. 

%O: intuition 
Intuitively, choosing the decision boundary based on test positivity rate is likely to yield the best performance. However, the test positivity rate is not available beforehand and tends to differ from the model score as shown in  Fig.~\ref{fig:teaser-results}(a).  More importantly,  the score estimation bias, i.e., difference between test positivity rate and the model score varies with uncertainty. Specifically, using Bayes rule, we observe that for input regions with a certain empirical train positivity rate, the “true positivity” (and hence test positivity rate) is shifted towards the global prior, with the shift being stronger for regions with low evidence. While Bayesian models try to adjust for this effect by combining the evidence, i.e., the observed train positivity  with  “model priors”,  there is still a significant bias when there is a mismatch between the model priors and the true prior in regions of weak evidence (high uncertainty).
%Intuitively, choosing the decision boundary based on empirical test positivity rate is likely to yield the best performance. However, the test positivity rate will not be available beforehand and tends to differ from the model score as shown in 
%Fig.~\ref{fig:teaser-results}(a). More importantly, the estimation bias, i.e., the difference between test positivity rate and the model score, varies with the uncertainty level. 
%This behavior arises from the fact that conditioned on the   observed train positivity rate taking a particular value, the likely “true positivity” for 
%While Bayesian models try to adjust for this effect by combining evidence with  “model priors”,  there is still a significant bias especially when model prior mismatches the true priors along with evidence is weak, i.e., for higher uncertainty.
%This behavior can be attributed to the fact that in Bayesian models, the model score combines both prior and evidence leading to a bias in the direction of the prior with the bias being higher when the evidence is weak, i.e., for higher  uncertainty.
Differential sampling across classes during training %(e.g., downsampling of negative class) 
further contributes to this bias. This finding that the same model score can map to different test positivity rates based on uncertainty levels indicates that the decision boundary chosen using score alone is likely to be suboptimal relative to the best one based on both
%prediction 
uncertainty and model score.
%This relationship between uncertainty and estimation bias indicates that the best decision boundary based on score alone  is likely to be sub-optimal relative to the best one based on both uncertainty and model score.
Fig. \ref{fig:teaser-results}(b) depicts maximum recall boundaries for a specified precision bound using score alone (yellow) and with both score and uncertainty estimates (red) validating this observation. 
 %that validate this observation. 

\noindent{\bf Contributions.} Below we summarize our contributions on leveraging the relationship between score estimation bias and uncertainty to improve classifier performance.

\noindent  1. Considering a Bayesian setting with Beta priors and  Posterior Network ~\citep{posterior_network} for uncertainty estimation, we analyse the behavior of test positivity rate and find that it depends on both score and uncertainty, and monotonically increases with score for a fixed uncertainty. There is also a dependence on the downsampling rate in case of differential sampling during training.
%the trend with respect to uncertainty depends on the sign of 1- \nu (s+\xi\nu -\omega) 
\\
\noindent  2. We introduce 2D decision boundary estimation problem in terms of maximizing recall for target precision (or vice versa). % using both uncertainty and score. %where the boundary is defined in terms of varying score thresholds for different ranges of uncertainty that are more aligned to the true empirical probabilities
%Keeping in view computational efficiency, we formulate the 2D-Binned Decision Boundary (2D-BDB) problem based on partitioning the two-dimensional model score x uncertainty space into bins.  We demonstrate that this problem is connected to bin-packing and prove that it is, in fact, NP-hard (for variable bin sizes) via a reduction from the subset-sum problem ~\citep{subsetsum}.\\
 Keeping in view computational efficiency, we partition the %two-dimensional 
model score $\times$ uncertainty space into bins and demonstrate that this is connected to bin-packing, and prove that it is NP-hard (for variable bin sizes) via reduction from the subset-sum problem ~\citep{subsetsum}.\\
\noindent 3.  We present multiple algorithms for solving the 2D binned decision boundary problem defined over score and uncertainty derived from any blackbox classification model. % For the general case of variable sized bins, we propose pseudo-polynomial time dynamic programming (DP) algorithms for computing the optimal 2D decision boundary.  
We propose an equi-weight bin construction by considering quantiles on uncertainty followed by further quantiles along scores. For this case,  we present a polynomial time DP algorithm that is guaranteed to be optimal. Additionally, we also propose a greedy algorithm that performs isotonic regression ~\citep{isotonicstout} independently for each uncertainty level, and selects a global threshold on calibrated probabilities.\\ %output by the isotonic regression.\\
\noindent 4. We present empirical results on three  datasets and demonstrate that our proposed %2D decision boundary 
algorithms yield 25\%-40\% gain in recall at high precision bounds over the vanilla thresholding based on score alone.  %Further, we observe that the greedy isotonic regression and the DP algorithms for equi-weight bins yield the best performance with reasonable computational effort. 

\section{Related Work}
\label{sec:related}
% 1. Uncertainty Estimation
% --- [some content in Sec 2.1 right now] 
% --- typical approaches, advances 
% --- Main point: In Sec 2, we chose to do theoretical exploration of connection bias based on posterior networks because of the analytic framework it provides. However, the contributions in Sec 3 and 4 are independent of the choice of uncertainty estimation and can be used with any uncertainty estimation method. 
% SM: Is the point of generality of Sec 3/4 coming through?

 \noindent{\bf Uncertainty Modeling.} Existing approaches for estimating uncertainty can be broadly categorized as Bayesian methods
 %bayesiannn
 \citep{bayesianlr,weightuncertainty,bayesiandlcv}, Monte Carlo methods \citep{mcdropout} and ensembles \citep{deepensembles}. Dropout and ensemble methods estimate uncertainty by sampling probability predictions from different sub-models during inference, and are compute intensive. Recently,  \citep{posterior_network} proposed Posterior Network that directly learns the posterior distribution over predicted probabilities, thus enabling fast uncertainty estimation for any input sample in a single forward pass  
 %Such an approach 
 and providing an efficient analytical framework for estimating both aleatoric and epistemic uncertainty.\\
% Such an approach
% and  providing an efficient analytical and computationally efficient framework for estimating both aleatoric and epistemic uncertainty.\\
%  In our current work, we  used Posterior network both for theoretical analysis and experiments.  However, our proposed decision boundary algorithms  generalize to  any uncertainty estimation approach.\\
% 2. Using Uncertainty for Decision Making
% -- papers that use these (medical domain + economics)
% -- however approaches are heuristic based or tied to specialized modeling
% -- our approach is generic and more principled 
\noindent{\bf Uncertainty-based Decision Making.} 
While there exists considerable work on using uncertainty along with model score to drive explore-exploit style online-learning approaches \citep{exploreexploituncertainty}, leveraging uncertainty to improve precision-recall performance %in a generic classification setting 
has not been rigorously explored in literature to the best of our knowledge. 
Approaches proposed in the domain of digital pathology either use heuristics to come up with a 2D-decision boundary defined in terms of both model score and estimated uncertainty \citep{heuristic2d2022}, or use simple uncertainty thresholds 
%as determined through cross-validation within training folds 
to isolate or abstain from generating predictions for low-confidence samples from test dataset, to boost model accuracy \citep{thresholding2022, ZHOU2022449}. \\
%In contrast, we propose a principled non-parametric approach with certain optimality properties and can be used with data from any domain. \\
% 3. Score calibration methods (e.g., histogram, isotonic regression)
% -- some stuff from the Nov-end writeup
% -- mention 2D-isotonic regression (but we didn't need to use it since  both theoretical and empirical analysis pointed to monotonicity not being strict) 
%-- While the isotonic regression within each uncertain level is greedy, the choice of global score threshold across levels allows for a better optimization. 
%-- [optional?] Currently we assume that binning configuration (#bins) is given but can use ideas from literature to jointly optimize the #bins and boundary
\noindent {\bf Model Score Recalibration.} These methods transform the model score into a well-calibrated probability using 
empirical observations on a hold-out set. Earlier approaches include histogram binning \citep{binningicml2001}, isotonic regression \citep{isotonicstout}, and temperature scaling \citep{calibrationmethods}, all of which consider the model score alone during recalibration. Uncertainty Toolbox \citep{chung2021uncertainty} implements recalibration methods taking into account both uncertainty and model score but is currently limited to regression. In our work, we propose an algorithm (MIST \ref{alg:algo7}) that first performs 1D-isotonic regression on samples within an uncertainty level to calibrate probabilities and then select a global threshold. In addition to achieving a superior decision boundary, this results in lower calibration error compared to using score alone.

\section{Relationship between Estimation Bias and Uncertainty}
\label{sec:analysis}

To understand the behavior of estimation bias, we consider a representative data generation scenario and analyse the dependence of estimation bias on uncertainty  with a focus on Posterior Network \footnote{Note that Theorem \ref{thm:bias}(a) on the relationship between train, true, and test positivity  is independent of the model choice. We consider Posterior Network only to express the bias in terms of model score and  uncertainty.}.
% though one can generalize it to other uncertainty modeling approaches. 

\noindent{\bf Notation.} Let $\x$ denote an input point and $y$ the corresponding target label that takes values from the set of class labels $\cC =\{ 0, 1 \}$ with $c$ denoting the index over the labels. See Appendix \ref{appendix:notations}. %Further, let  $\cD_{train}$ denote the data used for training, $\theta, \phi$ the model 
% parameters and $\z= \z(\x)$ the output representation from the model. 
We use $\pr(\cdot)$ to denote probability and $[i]_{lb}^{ub}$ to denote an index  iterating over integers in $\{lb, \cdots, ub\}$.

\noindent{\bf 3.1 Background: Posterior Network}
Posterior Network \citep{posterior_network}  estimates a closed-form posterior distribution over predicted class probabilities for any new input sample via density estimation as described  in Appendix \ref{appendix:posterior}.  For binary classification, the posterior distribution at  $\x$ is a Beta distribution with parameters estimated by combining the model prior with pseudo-counts generated based on the learned normalized densities and observed class counts. Denoting the model prior and observed counts for the class $c \in \cC$ by $\beta_c^{P}$ and $N_c$, the posterior distribution of predicted class probabilities %(a.k.a  epistemic distribution)
at $\x$ is given by
% \begin{wrapfigure}{r}{.3\textwidth}
% \begin{equation}
$q(\x)=  \mbox{Beta}( \alpha_1(\x),\alpha_0(\x) )$
% \end{equation}
% \end{wrapfigure}
where $\alpha_c(\x)= \beta_c^{P} + \beta_c(\x)$   and $\beta_c(\x) = N_c \pr(\z(\x) | c; \phi),~ \forall c  \in {\cal C}$. Here, $\z(\x)$ is the penultimate layer representation of $\x$ and $\phi$ denotes parameters of a normalizing flow.  Model score $S^{model}(\x)$ for positive class is given by 
% \begin{wrapfigure}{R}{.5\textwidth}
\begin{equation}
S^{model}(\x) =  \frac{ \beta_1^{P} +  \beta_1(\x)} { \sum_{c\in \cC} [\beta_c^{P} +  \beta_c(\x)]}  =  \frac{\alpha_1(\x)}{\alpha_1(\x) + \alpha_0(\x)}. 
\label{eqn:smodel}
\end{equation}
% \end{wrapfigure}
Uncertainty $u(\x)$ for $\x$ is given by differential entropy of distribution 
$H(q(\x))$\footnote{$H(q(\x))=  \log \cB (\alpha_0,\alpha_1) - (\alpha_0 + \alpha_1-2)\psi(\alpha_0+\alpha_1) - \sum_{c \in \cC} (\alpha_c - 1)\psi(\alpha_c)$ where 
$\psi(\cdot)$ is the digamma function and $\cB(\cdot,\cdot)$ is the Beta function.}.  Since $q(\x)$ is Beta distribution, for same score,(i.e., $\alpha_1(\x)$/$\alpha_0(\x)$), uncertainty is higher when $\sum_{c \in \cC} \alpha_c(\x)$ is lower.

\begin{wrapfigure}{R}{0.4\textwidth}
%\vskip 0.2in
\vspace{-3.5mm}
\begin{center}
\centerline{\includegraphics[width=0.35\columnwidth,height=3cm]{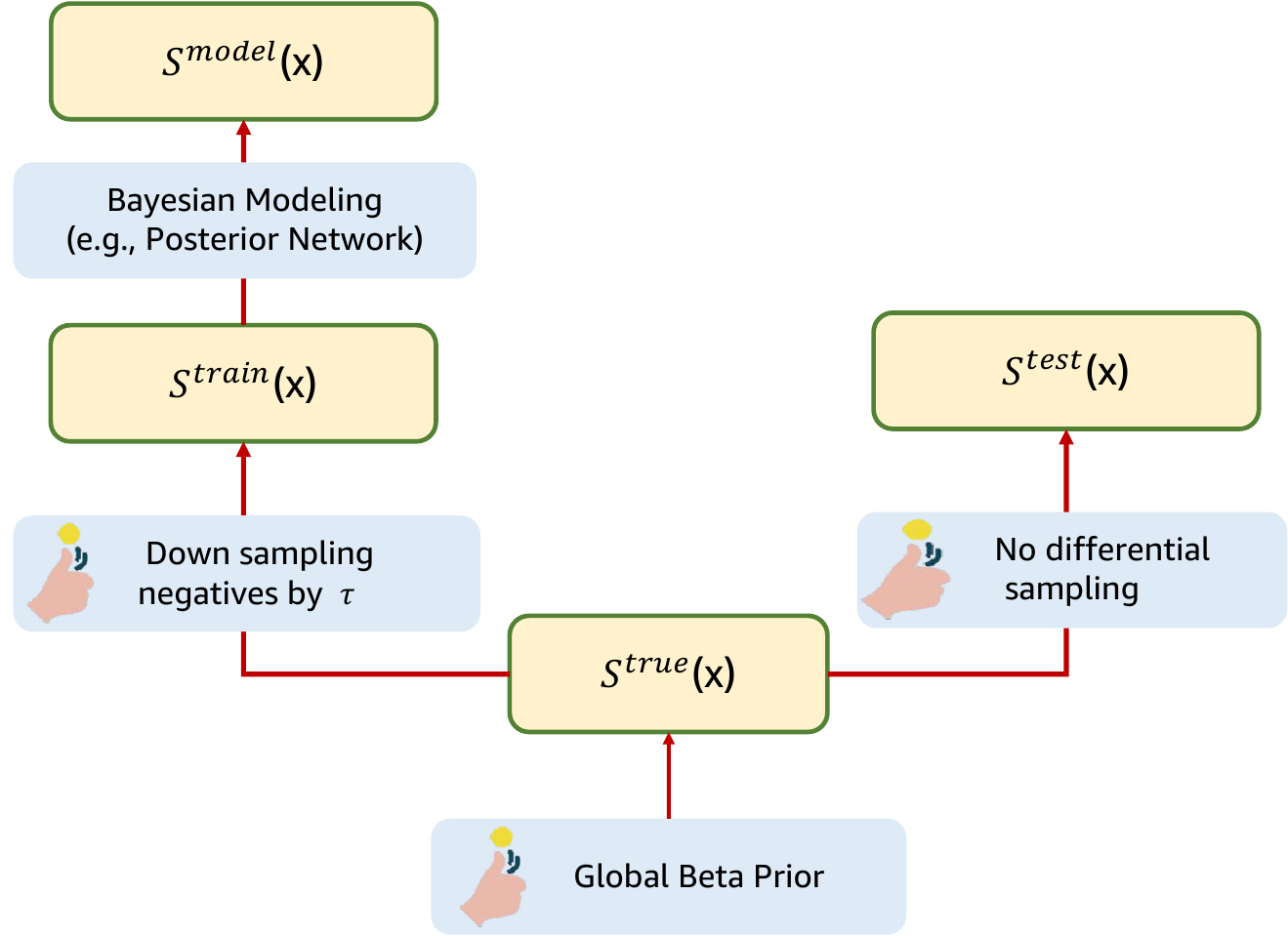}}
\caption{\footnotesize Dependencies among various positivity rates and the model score.}
\vspace{-3.5mm}
\label{fig:dependencies}
\end{center}
%\setlength{\belowcaptionskip}{-10mm}
%\vskip -0.2in

\end{wrapfigure}

\noindent{\bf 3.2 Analysis of Estimation Bias}: For an input point $\x$, let $S^{true}(\x), S^{train}(\x),  S^{test}(\x),$ and $S^{model}(\x)$ denote the true positivity, empirical positivity in the train and test sets, and the model score respectively.  Assuming the train and test sets are drawn from the same underlying distribution with possible differential sampling across the classes, these variables are dependent on each other as shown in Fig. \ref{fig:dependencies}.  We consider the following generation mechanism where the true positivity rate is sampled from a global beta prior, i.e., 
$S^{true}(\x) \sim Beta(\beta_1^T, \beta_0^T) $.  The labels $y(\x)$ in the test set are generated via Bernoulli distribution centered at $S^{true}(\x)$. In the case of train set, we assume that the negative class is sampled at rate  $\frac{1}{\tau}$ compared to positive class. Note that $\tau >1 $  corresponds to undersampled negatives while $\tau <1$ corresponds to oversampled negative class. We define $\gamma(\x) = \frac{\beta_1^{P} + \beta_0^{P}} {\beta_1(\x) + \beta_0(\x)}$, i.e., the ratio of combined priors to the combined likelihood evidence.
%i.e., labels can be viewed to be generated from a Bernoulli centered around $\frac{\tau S^{true}(x)} {\tau S^{true}(x) + (1-S^{true}(x)) }$. 

Given $S^{model}(\x)$ of Posterior Network and $\gamma(\x)$, the train positivity rate is fixed (Lemma \ref{thm:model_train}). Using Bayes rule, one can then estimate expected true and test positivity rate conditioned on train positivity (or equivalently model score) as in Theorem \ref{thm:bias} (Proof details in Appendix \ref{appendix:new_analysis}).

\begin{theorem} 
\label{thm:bias}
When data is generated as per Fig. ~\ref{fig:dependencies} and negative class is  undersampled at the rate $\frac{1}{\tau}$:\\
%the following results hold:\\
(a) The expected test and true positivity rate conditioned on the train positivity  are equal and correspond to the expectation of the distribution,  
$$ Q(r)  = \frac{C}{(1+(\tau-1) r)^n} Beta(n(\xi\lambda(\x) + S^{train}(\x)),n( (1-\xi)\lambda(\x) + 1 - S^{train}(\x) )). $$

When there is no differential sampling, i.e., $\tau=1$, the expectation has a closed form and is given by
$$ E [ S^{true}(\x) | S^{train}(\x) ] = E [ S^{test}(\x) | S^{train}(\x)] = \frac{S^{train}(\x) + \xi\lambda(\x) }{1 + \lambda(\x)}.$$
%\begin{itemize}
%\item 
Here, $n =n(\x)= \beta_1(\x) + \beta_0(\x)$ denotes evidence, 
~$C$ is a normalizing constant, ~$\xi =\frac{\beta_1^{T}}{\beta_1^{T} + \beta_0^{T}}$ is the positive global prior, and $\lambda(\x) = \frac{\beta_1^{T} + \beta_0^{T}} {\beta_1(\x) + \beta_0(\x)}$ is  the ratio of global priors to evidence.
%\item  $\omega =\frac{\beta_1^{P}}{\beta_1^{P} + \beta_0^{P}}$ and $\xi =\frac{\beta_1^{T}}{\beta_1^{T} + \beta_0^{T}}$  are the positive prior ratios,  
%\item $\nu = \frac{\beta_1^{T} + \beta_0^{T}} {\beta_1^P + \beta_0^P}$ is the ratio of global and model priors.
%_{\text{true-priors : model-priors}}$ 
%\item $\nu = \underbrace{ \frac{\beta_1^{T} + \beta_0^{T}} {\beta_1^P + \beta_0^P}}$ is the ratio of global and model priors.
%_{\text{true-priors : model-priors}}$ 
\\
%\end{itemize}
(b) For Posterior Networks, test and true positivity rate conditioned on model score $S^{model}(\x)$ can be obtained using $S^{train}(\x) = S^{model}(\x) - (\omega - S^{model}(\x)) \gamma(\x)$. For $\tau =1$, the estimation bias, i.e. difference between model score and test positivity is given by $\frac{(S^{model}(\x)(\nu-1) +\omega -\xi\nu)\gamma(\x)}{1+\nu\gamma(\x)},$
where 
%\begin{itemize}
%\item 
    $\omega =\frac{\beta_1^{P}}{\beta_1^{P} + \beta_0^{P}}$ %is the positive model prior, 
    and  $\nu = \frac{\lambda(\x)}{\gamma(\x)} =\frac{\beta_1^{T} + \beta_0^{T}} {\beta_1^P + \beta_0^P}$ is the ratio of global and model priors.
%_{\text{true-priors : model-priors}}$ 
%\end{itemize}

%For $\tau =1$, the estimation bias, i.e. difference between model score ($S^{model}(x) =s(x)$) and test positivity is given by $\frac{(s(x)(1-\nu) + \xi\nu -\omega)\gamma(x)}{1+\nu\gamma(\x)}.$ 

%$$Q(r) =  \frac{C}{(1+(\tau-1) r)^n}  Beta(n(s - (s-w-\xi\nu) \gamma(x) ), n ( 1+ \nu \gamma(x) - s + (s-\omega - \xi\nu)\gamma(x) ) ).$$
%For $\tau =1$, this reduces to 
%$$ E [ S^{true}(x) | S^{model}(x) = s ] = E [ S^{test}(x) | S^{model}(x) = s ] = \frac{s  -(s - \xi\nu -\omega) \gamma(\x)}{1+\nu\gamma(\x)}. $$ 
\end{theorem}

\noindent {\bf Interpretation of $\gamma(\x)$.}  Note that $\sum_c {\alpha_c(\x)}  =  [\sum_c \beta^{P}_c ] (1 + \frac{1}{\gamma(\x)})$.   For a fixed score,  $\sum_c{\alpha_c(\x)}$  varies  inversely with uncertainty $u(\x)=H(q(\x))$,
making the latter positively correlated with
%implying  the latter is positively correlated with 
$\gamma(\x)$. 
%is positively correlated with $H(q(\x))$.

\noindent  {\bf No differential sampling $(\tau =1)$}.
%In this case, expected test positivity equals $\frac{s + (s + \xi\nu - \omega)\gamma(x)}{ 1 + \nu\gamma(x)}.$ 
Since the model scores are estimated by combining the model priors and the evidence,  $S^{model}(\x)=s(\x)$  differs from the train positivity rate in the direction of the model prior ratio $\omega$. On the other hand, expected true and test positivity rate differ from train positivity rate in the direction of true class prior ratio $\xi$. When the model prior matches true class prior both on positive class ratio and magnitude, $i.e., \nu=1, \xi=\omega$,  there is no estimation bias. In practice, model priors are often chosen to have low magnitude and estimation bias is primarily influenced by global prior ratio with  overestimation (i.e., expected test positivity $<$ model score) in the higher score range $(\xi < s(\x))$ and the opposite is true when $(\xi > s(\x))$.  The extent of bias depends on relative strengths of priors w.r.t evidence denoted by $\gamma(\x)$, which is correlated with uncertainty. For this case, the expected test positivity is linear and monotonically increasing in model score. The trend with respect to uncertainty depends on sign of $(s(\x)(\nu-1) +\omega -\xi\nu)$.

\begin{wrapfigure}{l}{0.48\textwidth}
\hfill
\vspace{-4mm}
\subfigure[]{\includegraphics[width=0.48\linewidth,height=2.8cm]{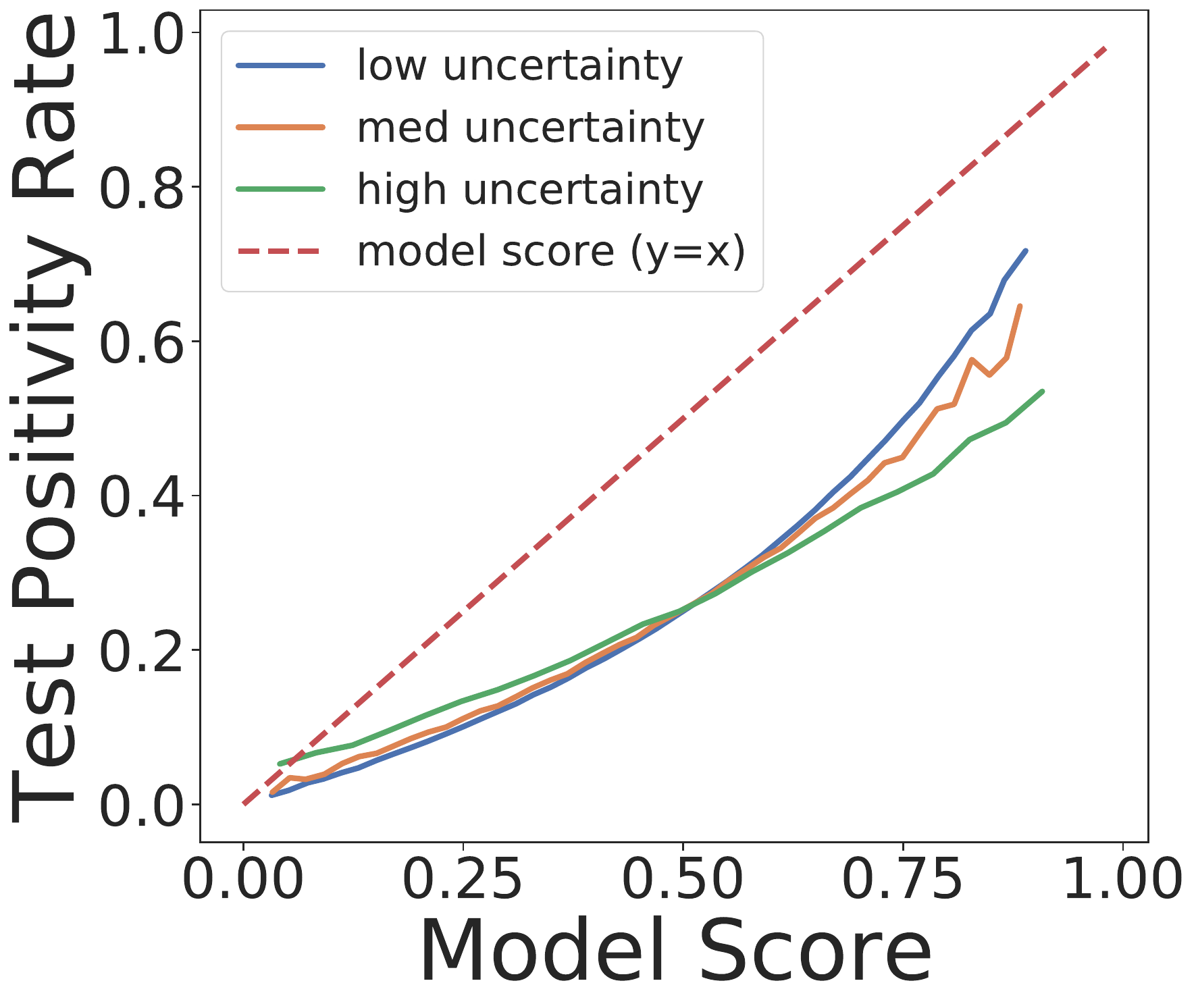}}
\hfill
\subfigure[]{\includegraphics[width=0.48\linewidth,height=2.8cm]{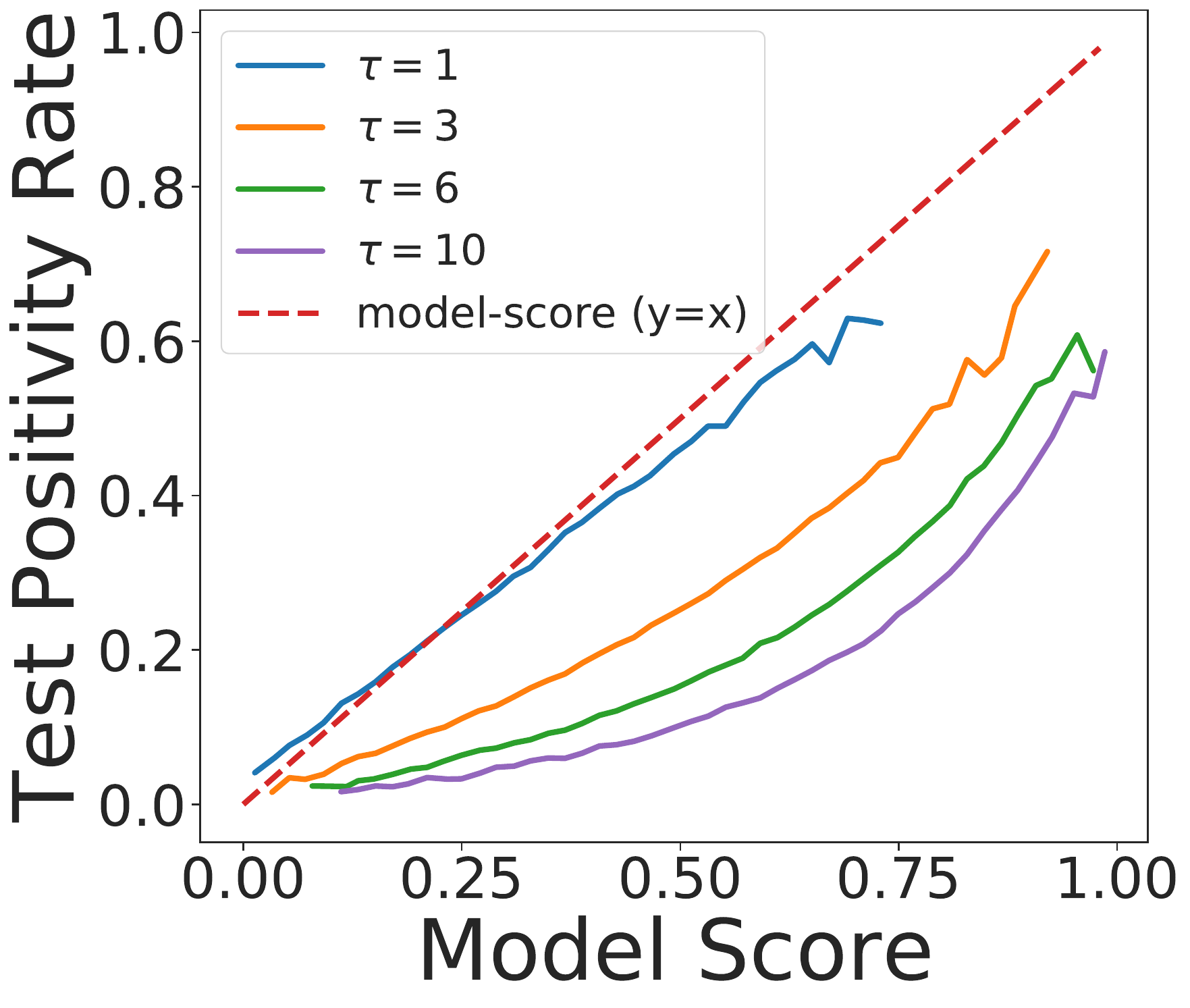}}
\hfill
% \vspace{-3mm}
\caption{\footnotesize  Test positivity vs. model score curves for (a) few choices of $\gamma(\x)$ with $\omega=0.5,~\tau=3$, and (b) few values of $\tau$ with
$\omega=0.5$ and medium uncertainty using data simulation as per Fig. \ref{fig:dependencies}.}
\label{fig:biasanalysis}
\vspace{-6mm}
\end{wrapfigure}

\noindent  {\bf General case $(\tau > 1)$.} Here, the expected behavior is affected not only by the interplay of the model prior, true class prior and evidence as in case of $\tau=1$, but also the differential sampling.  While the first aspect is similar to the case $\tau=1$, %and results in over and under estimation in different score regions, 
the second aspect results in  overestimation across the entire score range with the extent of bias increasing with $\tau$.  Fig. ~\ref{fig:biasanalysis}(a) shows the expected positivity rate for a few different choices of $\gamma(\x)$ and a fixed choice of $\omega=0.5$ and $\tau=10$ while Fig.~\ref{fig:biasanalysis}(b) shows the variation with different choices of $\tau$. We validate this behavior by comparison with empirical observations in Sec. \ref{sec:experiment}.
% and Appendix \ref{appendix:experiment}.
%SM: More details on the trends in the Appendix [Optional for AMLC]

The primary takeaway from Theorem \ref{thm:bias} is that the score estimation bias depends on score and uncertainty. For a given model score, different samples can correspond to different true positivity rates based on uncertainty level, opening an opportunity to improve the quality of the decision boundary by considering both score and uncertainty. However, a direct adjustment of model score based on Theorem \ref{thm:bias} is not feasible or effective since the actual prior and precise nature of distributional difference between test and train settings might not be known. 
% This is fairly common in large-scale real-life applications where legacy blackbox models are used for downstream prediction tasks or when there is  temporal drift. 
Further, even when there is information on differential sampling rate used in training, class-conditional densities learned from sampled distributions tend to be different from original distribution especially over sparse regions.

\section{2-D Decision Boundary Problem} 
\label{sec:prob}
%O: General approach; reference figure
%O: problem formulation
%SM: Do we need to define recall and precision (isn't that standard) - we could use TP,FP notation 

\begin{wrapfigure}{L}{0.5\textwidth}
%\vskip 0.2in
\begin{center}
\centerline{\includegraphics[width=0.48\columnwidth,height=3.5cm]{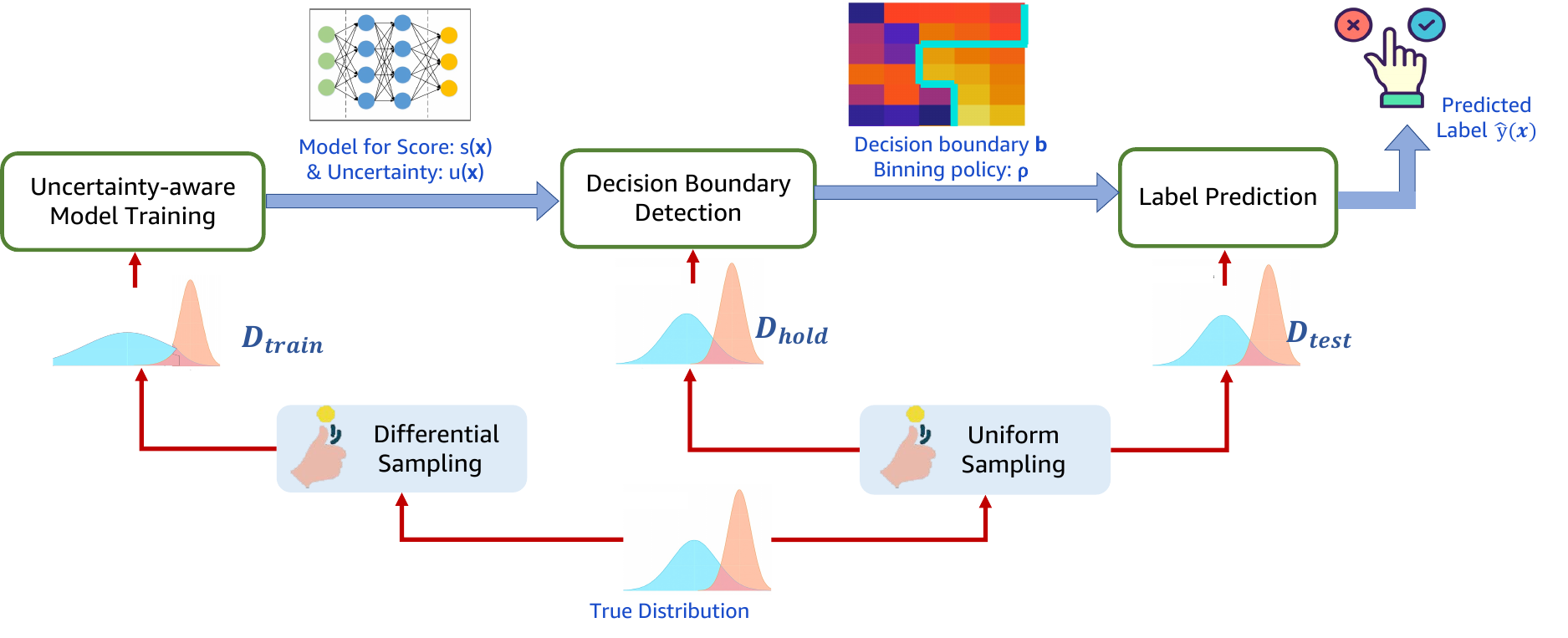}}
\caption{\footnotesize Binary classification with model training followed by decision boundary selection on hold-out set.}
\label{fig:schematic}
\end{center}
%\setlength{\belowcaptionskip}{-10mm}
%\vskip -0.2in
\vspace{-10mm}
\end{wrapfigure}

Given an input space $\cX$ and binary labels $\cC$, binary classification typically involves finding a mapping $\psi_{\myb}:\cX \rightarrow \cC$ that optimizes application-specific performance. Fig.~\ref{fig:schematic} depicts a typical supervised learning setting where labeled training data $\cD_{train}$, created via differential sampling along classes, is used to learn a model. A hold-out labeled set $\cD_{hold}$, disjoint from training and similar in distribution to deployment setting is used to construct a labeling function $\psi_{\myb}$ based on model output. 

%O: simple 1 D boundary, 
%O: 2D case - monotonicity wrt score /generalisation and continuous curve
Traditionally, we use the labeling function $\psi_{\myb}(\x) = \mathbbm{1}[s(\x) \geq b]$ with boundary $\myb=[b]$  defined in terms of a score threshold optimized based on the hold-out set $\cD_{hold}$. When the model outputs both score and uncertainty $(s(\x), u(\x))$, 
we have a 2D space to be partitioned into positive and negative regions. In Sec. \ref{sec:analysis}, we observed that the true positivity rate is monotonic with respect to score for a fixed uncertainty.
Hence, we consider a boundary of the form $\psi_{\myb}(\x) = \mathbbm{1} [s(\x) \geq b(u(\x)) ]$, where $b(u)$ is score threshold for uncertainty $u$. 

%Traditionally, %the model output is a single score $s(\x)$ and we use the labeling function $\psi_{\myb}(\x) = \mathbbm{1}[s(\x) \geq b]$ with boundary $\myb=[b]$  defined in terms of a scalar score threshold that is  optimized based on the hold-out set $\cD_{hold}$. When the model outputs both score and uncertainty $(s(\x), u(\x))$, for an input $\x$, 
%we have a 2D space to be partitioned into positive and negative regions. In Section \ref{sec:analysis}, we observed that the true positivity rate is monotonic with respect to score for a fixed uncertainty level.
%Hence, to ensure generalisation from the hold-out set to the deployment setting, 
%Hence, we consider a boundary of the form $\psi_{\myb}(x) = \mathbbm{1} [s(\x) \geq b(u(\x)) ]$, where $b(u)$ is score threshold when uncertainty is $u$. 

%O:  binning-motivation and binning-notation
To ensure tractability of decision boundary selection, a natural approach is to either limit $\myb$ to 
a specific parametric family or discretize the uncertainty levels. We prefer the latter option as it allows generalization to multiple uncertainty estimation methods. Specifically, we partition the 2D
score-uncertainty space into bins forming a grid such that the binning preserves the ordering over the space. 
(i.e., lower values go to lower level bins). 
This binning could be via independent splitting on both dimensions, or by partitioning on one dimension followed by a nested splitting on the other.
%This binning can be obtained using different methods, e.g., by independently splitting on both dimensions, or by partitioning along one dimension followed by a nested partitioning on the other. 

Let $\cS$ and $\cU$ denote the possible range of  score and uncertainty values, respectively. Assuming $K$ and $L$ denote the desired number of uncertainty and score bins, let 
$\rho: \cU \times \cS \mapsto \{1,\cdots, K\} \times \{1,\cdots, L\}$ denote a partitioning such that any 
score-uncertainty pair $(u, s)$ is mapped to a unique bin $(i, j) = (\rho^U(u), \rho^S(s))$ in the $K \times L$ grid. 
We capture relevant information from the hold-out set via two $K \times L$ matrices $[p(i,j)]$ and $[n(i,j)]$ where $p(i,j)$ and $n(i,j)$ denote the positive and the total number of samples in the hold-out set mapped to the bin $(i,j)$ in the grid. Using this grid representation, we now define the 2D Binned Decision Boundary problem. For concreteness, we focus on maximizing recall subject to a precision bound though our results can be generalized to other settings where the optimal operating point can be derived from the PR curve.

%O: problem definition
\noindent {\bf 2D Binned Decision Boundary Problem (2D-BDB):}
Given a  $K \times L$ grid of bins with positive sample counts $[p(i,j) ]_{K \times L}$ and total sample counts  $[n(i,j) ]_{K \times L}$ corresponding to the hold-out set $D_{hold}$ and a desired precision bound $\sigma$, find the optimal boundary  $\myb = [b(i)]_{i=1}^K$ that maximizes recall subject to the precision bound as shown in Eqn. \ref{eqn:maxrecallforprecision}.
\begin{wrapfigure}{r}{.43\textwidth}
\vspace{-3mm}
\begin{equation}
\boxed{\underset{\myb ~~
\text{s.t.} ~~precision(\psi_{\myb}) \geq \sigma; ~ 0 \leq b[i] \leq L }{\text{argmax}} recall (\psi_{\myb})}
\label{eqn:maxrecallforprecision}
\end{equation}
\vspace{-8mm}
\end{wrapfigure}
Here $recall(\psi_{\myb})$  and $precision(\psi_{\myb})$ denote the recall and precision
%, respectively, 
of the labeling function $\psi_{b}(\x) = \mathbbm{1} [\rho^{\cS}(s(\x)) > b (\rho^{\cU}(u(\x))) ]$ with respect to true labels in $D_{hold}$. %~\citep{standard-classification-metrics}. $\Box$
%SM: do we need to define recall and precision ?
While $D_{hold}$ is used to determine the optimal boundary, actual efficacy is determined by the  performance on unseen test data.

%SM: ADD to footnote if necessary
%Note $b(i)=L$ corresponds to the case of not selecting any point from the $i^{th}$ uncertainty level as positive.

\noindent{\bf Connection to Knapsack Problem.} Note that the 2D decision boundary problem has similarities to the knapsack problem in the sense that given a set of items (i.e., bins), we are interested in choosing a subset that maximizes a certain “profit” aspect while adhering to a bound on a specific “cost” aspect.  However, there are two key differences - 1) 
the knapsack problem has notions of cost and profit, while in our case we have precision and
recall. 
% Recall is similar to profit, in the sense that it is additive. 
On the other hand, our cost aspect is the false discovery rate (i.e., 1- precision) which is not additive, and the change in precision  due to selection of a bin depends on previously selected bins, and 2) our problem setting has more structure since bins are arranged in a 2D-space with constraints on how these are selected.

\section{2-D Decision Boundary  Algorithms} 
\label{sec:algo}
 We provide results of computational complexity of 2D-BDB problem along with various solutions. 
% We present various solutions to 2D-BDB problem along with computational complexity. 
\subsection{NP-hardness Result}
It turns out the  problem of computing the optimal decision boundary over a 2D grid of bins (2D-BDB) is intractable for the general case where the bins have different  sizes. We use a reduction from NP-hard subset-sum problem ~\citep{npcompleteness} for the proof, detailed in Appendix \ref{appendix:nphard}.
\begin{theorem} The problem of computing an optimal 2D-binned decision boundary is NP-hard. 
$\Box$
\label{thm:nphard}
\end{theorem}
\makeatother

\begin{wrapfigure}{R}{0.5\textwidth}
\vspace{-8mm}
    \begin{minipage}{0.5\textwidth}
      \begin{algorithm}[H]\small
      \caption{Optimal Equi-weight DP-based Multi-Thresholds [EW-DPMT]}
 \label{alg:algo3}
 \begin{algorithmic}
  \STATE \hspace*{-4mm} {\bf Input:} Equi-sized $K \times L$ grid with positive sample counts $[p(i,j) ]_{K \times L}$, total count $N$, precision bound $\sigma$
  \STATE \hspace*{-5mm} {\bf Output:} maximum (unnormalized) recall $R^*$ and corresponding optimal boundary $\myb^*$ for precision $\geq \sigma$
 
  \STATE \textit{// Initialization}
  \STATE $R(i,m) =  -\infty$; $b(i,m,i’) =  -1$;  ~$ \forall [i]_1^K, ~[i']_1^K, ~ [m]_0^{KL} $ 
  \STATE \textit{// Pre-computation of cumulative sums of positives}
  \STATE $\pi(i,0) = 0, ~~~ [i]_1^K $ 
  \STATE $\pi(i,j) = \sum_{j'=L-j+1}^{L} p(i,j'), ~~ [i]_1^K, ~[j]_1^L $
  \STATE \textit{// Base Case: First Uncertainty Level}
  \STATE $R(1,m) = \pi(1,m)$; $b(1,m,1) = L-m,~~~ [m]_0^L$
 \STATE \textit{// Decomposition: Higher Uncertainty Levels}
 \FOR{$i=2$ to $K$}
  \FOR{$m=0$ to $iL$}
  \STATE $j^* = \underset{0 \leq j \leq L}{\text{argmax}} [ \pi(i,j) +  R(i-1, m-j) ]$
  \STATE $R(i, m) = \pi(i,j^*) + R(i-1, m-j^*)$
  \STATE $b(i,m,:) = b(i-1,m-j^*,:)$  
  \STATE $b(i,m,i) = L- j^* $
    \ENDFOR
    \ENDFOR
\STATE \textit{// Maximum Recall for Precision}
\STATE $m^* =  \underset{0 \leq m \leq KL ~s.t. \frac{KL}{mN} R(K,m) \geq \sigma } {\text{argmax}}  [ R(K,m)] $
\STATE $R^* = R(K,m^*)$; $\myb^* = b(K,m^*,)$
\STATE {\bf return} $(R^*, \myb^*)$
\end{algorithmic}
\end{algorithm}
\end{minipage}
\vspace{-5mm}
\end{wrapfigure}
\setlength{\textfloatsep}{0.2cm}
\setlength{\floatsep}{0.2cm}

\subsection{Equi-weight Binning Case} 
A primary reason for the intractability of 2D-BDB  problem is that one cannot ascertain the relative “goodness” (i.e., recall subject to precision bound) of a pair of bins based on their positivity rates alone. For instance, it is possible that a bin with lower positivity rate  might be preferable to one with higher positivity rate due to different number of samples. To address this, we propose a binning policy that preserves the partial ordering along score and uncertainty yielding equal-sized bins. We design an optimal algorithm for this special case using the fact that a bin with higher positivity is preferable among two bins of the same size.

\noindent {\bf Binning strategy}: To construct an equi-weight $ K \times L$ grid, we first partition the samples in $D_{hold}$ into $K$ quantiles along the uncertainty dimension and then split each of these $K$ quantiles into $L$ quantiles along the score dimension. %~\footnote{A similar scheme with first split along score and then along uncertainty is possible but skipped for brevity.}
% This policy yields bins that have $ \simeq \frac{1} {KL}$ fraction of the entire sample size. 
The bin indexed by $(i, j)$ contains samples from $i^{th}$ global uncertainty quantile and the $j^{th}$ score quantile local to $i^{th}$ uncertainty quantile. This mapping preserves the partial ordering that for any given score level, the uncertainty bin indices are monotonic with respect to its actual values. 
% Similarly, for any given uncertainty level, the score bin indices are monotonic with respect to the score values. 
Note that while this binning yields equal-sized bins on $D_{hold}$, using same boundaries on the similarly distributed test set will only yield approximately equal bins.

\noindent {\bf Dynamic Programming (DP) Algorithm}: 
For equi-weight binning, we propose a DP algorithm (Algorithm \ref{alg:algo3}) for the 2D-BDB problem that identifies a maximum recall decision boundary for a given precision bound by  constructing possible boundaries over increasing uncertainty levels. For $1 \leq i \leq K, 0 \leq m \leq KL$, let  $R(i,m)$  denote the maximum true positives for any boundary over the sub-grid with uncertainty levels upto the $i^{th}$ uncertainty level such that the boundary has exactly $m$ bins in its positive region. Further, let $b(i,m,:)$  denote the optimal boundary that achieves this maximum with $b(i,m,i')$ denoting the boundary position for the uncertainty level $i' (\leq i)$. Since bins are  equi-sized, for a fixed positive bin count, the set with most  positives yields the highest precision and recall. %relative to any other choice. 
For the base case $i=1$, feasible solution exists only for $0 \leq m \leq L$ and corresponds to picking exactly $m$ bins, i.e., score threshold index $b(1,m,1) = L - m$.
For $i>1$, we can decompose the estimation of maximum recall as follows. Let $j$ be the number of positive region bins from the $i^{th}$ uncertainty level. Then the budget available for the lower $(i-1)$ uncertainty levels is exactly $m-j$. Hence, we have, 
$R(i, m) = \underset{0 \leq j \leq L}{\text{max}} [\pi(i,j)  +  R(i-1, m-j)]$, where $\pi(i,j)= \sum_{j'=L-j+1}^{L} p(i,j')$, i.e., the count of positives in the $j$ highest score bins. The optimal boundary $b(i,m,:)$ is obtained by setting  $b(i,m,i) = L- j^*$ and the remaining thresholds to that of $b(i-1, m-j^*, :)$  where $j^*$ is the optimal choice of $j$ in the above recursion. Performing this computation progressively for all uncertainty levels and positive bin budgets yields maximum recall over the entire grid for each choice of bin budget. This is equivalent to obtaining the entire PR curve and permits choosing the optimal solution for a given precision bound. From $R(K, m)$,  we can choose the largest $m$ that meets the desired input precision bound to achieve optimal recall. The overall computation time complexity is $O(K^2L^2)$. More details in Appendix \ref{appendix:algorithm}.

\subsection{Other Algorithms}  
Even though the 2D-BDB problem with variable sized bins is NP-hard, it permits an optimal pseudo-polynomial time DP solution similar to the one presented above. \textsc{Variable-Weight DP based Multi-Thresholds (VW-DPMT))} (\ref{alg:algo5}) tracks best recall at sample level instead of bin-level as in \textsc{EW-DPMT} (\ref{alg:algo3}). We also consider two greedy algorithms that have lower computational complexity than the DP solution, and are applicable to both variable sized and equal-size bins. The first, \textsc{Greedy-Multi-Threshold (GMT)}, computes score thresholds that maximize recall for the given precision bound independently 
for each uncertainty level. 
% This has a time complexity of $O(KL)$. 
The second algorithm \textsc{Multi-Isotonic-Single Threshold (MIST)} is based on recalibrating scores within each uncertainty level independently using 1-D isotonic regression. 
%Here, 
We identify a global threshold on calibrated probabilities that maximizes recall over the entire grid so that the precision bound is satisfied.  Since the recalibrated scores are monotonic with respect to model score,  the global threshold maps to distinct score quantile indices for each uncertainty level. This has a time complexity of $O(KL\log(KL))$.

\section{Empirical Evaluation}
\label{sec:experiment}
We investigate the impact of leveraging uncertainty estimates along with the model score for decision-making with focus on the research questions listed in Sec. \ref{sec:intro}.
\subsection{Experimental Setup}
\label{subsec:exp_setup}
\textbf{Datasets}: For evaluation, we use three binary classification datasets: 
(i) \texttt{Criteo}: 
%An imbalanced dataset for CTR prediction
An online advertising dataset consisting of $\sim$45 MM ad impressions with click outcomes, each with
%served in 7 day period. Each sample has
13 continuous and 26 categorical features. We use the  %standard 
split of $72\%:18\%:10\%$ for train-validation-test from the \href{https://github.com/openbenchmark/BARS/tree/main/datasets/Criteo#Criteo_x1} {benchmark}, 
%{website}.
(ii) \texttt{Avazu}: Another CTR prediction dataset comprising  $\sim$40 MM samples each with 22 features describing user and ad attributes. 
We use the train-validation-test splits  of $70\%:10\%:20\%$, from the 
%benchmark
\href{https://github.com/openbenchmark/BARS/tree/main/datasets/Avazu#avazu_x1}{benchmark},
%{website}.
(iii) \texttt{E-Com}: A proprietary e-commerce dataset with $\sim$4 MM samples where the positive class indicates a rare user action. We create train-validation-test sets in the proportion $50\%:12\%:38\%$ from different time periods. In all cases, we train with varying degrees of undersampling of negative class with test set as in the original distribution.

\textbf{Training}: For \texttt{Criteo} and \texttt{Avazu}, we use the \href{https://github.com/openbenchmark/BARS/tree/main/ctr_prediction/benchmarks/SAM/SAM_avazu_x1}{SAM architecture} 
% \hyperref[SAM architecture]{https://github.com/openbenchmark/BARS/tree/main/ctr_prediction/benchmarks/SAM/SAM_avazu_x1}
\citep{SAM_architecture} as the backbone with 1 fully-connected layer and 6 radial flow layers for class distribution estimation. 
%as per Posterior Networks. 
%Validation set is used to identify model with best PR-AUC. 
For \texttt{E-Com}, we trained a FT-Transformer \citep{gorishniy2021revisiting} 
%based 
backbone with 8 radial flow layers.\\
%optimized for high PR-AUC on hold-out dataset.\\
\textbf{Binning strategies}: We consider two options: (i) \texttt{Equi-span} where the uncertainty and score ranges are divided into equal sized $K$ and $L$ intervals, respectively. Samples with uncertainty in the $i^{th}$ uncertainty interval, and score in the $j^{th}$ score interval are mapped to bin $(i,j)$. 
%Note that this is a special case of variable sized binning; 
(ii) \texttt{Equi-weight} where we first partition along uncertainty and then score as described in Sec. \ref{sec:algo}.\\
\textbf{Algorithms}: We compare our proposed  decision boundary selection methods against (i) the baseline of using only score, \textsc{Single Threshold (ST)} disregarding uncertainty, and (ii) a state-of-the-art 2D decision boundary detection method for medical diagnosis~\citep{heuristic2d2022}, which we call \textsc{Heuristic Recalibration (HR)}. The greedy algorithms (\textsc{GMT}, \textsc{MIST}), variable weight DP algorithm (\textsc{VW-DPMT}) are evaluated on both \texttt{Equi-weight} and \texttt{Equi-span} settings, and the equi-weight DP algorithm (\textsc{EW-DPMT}) only on the former. All results are on the test sets.
%unless mentioned otherwise. 
%\textbf{Metrics}: We evaluate the performance of the various algorithms in terms of the area under the PR curve both for the entire range (PR-AUC).
% and separately for the high precision range($\geq XX$)(PR-AUC-XX). 
%along with recall at a high precision bound (XX).
\subsection{Results and Discussion}
\label{subsec:results}
\noindent{\bf RQ1: Estimation Bias Dependence on Score \& Uncertainty.} From Sec. \ref{sec:analysis}, we observe that the estimation bias and thus the test positivity rate is dependent on both uncertainty and the model score. Fig. \ref{fig:teaser-results} and Fig. \ref{fig:biasanalysis} show the empirically observed behavior on the \texttt{Criteo} dataset and synthetic data generated as per Fig. \ref{fig:dependencies} respectively with  $\omega=0.5,~\tau=3,\xi =0.25$ in both cases. The observed empirical trends are broadly aligned with the theoretical expectations in Sec. \ref{sec:analysis} even though the assumption of a global Beta prior might not be perfectly valid.
%While the empirical observations are not exactly in line with the theoretical expectations since the assumptions of a global Beta prior in Section \ref{sec:analysis} might not be perfectly valid, the broader trends hold.
%~\footnote{Fig. ~\ref{fig:teaser-results} shows differential behavior w.r.t. uncertainty for the same score on \texttt{Criteo} dataset}.% but overestimation over a much larger score range.}.
In particular, the separation between uncertainty levels is more prominent for the higher score range in these imbalanced datasets, pointing to the criticality of considering uncertainty for applications where high precision is desirable. 
%%%%%%%%%%%% SM:EDIT THIS
To validate this further, we examine subsets of data  where the algorithms \textsc{EW-DPMT} and \textsc{ST}  differ on the  decision boundary  for 90\% precision (with \#score-bins = 500, \#uncertainty-bins = 3) on \texttt{Criteo} dataset. We observe that the bin $[(s(\x), u(\x))=(0.984, 0)]$ with positivity rate $0.91$  is labeled as positive by \textsc{EW-DPMT} but negative by  \textsc{ST} while the reverse is true for the bin $[(s(\x), u(\x))=(0.996, 0.667)]$ with a positivity rate $0.87$. Note that $((s(\x), u(\x))$ are percentiles here. This variation of positivity with uncertainty for the same score substantiates the benefits of flexible 2D decision boundary estimation. More analysis of these bins in Appendix  \ref{appendix:incremental_pos}.

\begin{table*}[tp]%[!htbp]
% \vspace{-2mm}
% \begin{minipage}[b]{0.6\textwidth}
\begin{center}
\begin{small}
\begin{tabular}{|c||c|c||c|c||c|c|}
\hline
  & \multicolumn{2}{|c||}{\texttt{Criteo}, 90\% Precision} & \multicolumn{2}{|c||}{\texttt{Avazu}, 70\% Precision} & \multicolumn{2}{|c|}{\texttt{E-Com}, 70\% Precision} \\
%  & \multicolumn{2}{|c|}{CIFAR10-bin} \\
 \hline
 \hline
 & \multicolumn{2}{|c||}{\small{$\tau$=3, Pos:Neg = 1:3}} & \multicolumn{2}{|c||}{\small{$\tau$=5, Pos:Neg = 1:5}}& \multicolumn{2}{|c|}{\small{$\tau$=5, Pos:Neg = 1:24}} \\
\hline

\hline
 \textbf{Algorithm} & \textbf{Equi-Span} & \textbf{Equi-weight} & \textbf{Equi-Span} & \textbf{Equi-weight} & \textbf{Equi-Span} & \textbf{Equi-weight} \\
% & Equi-Span & Equi-weight \\
\hline
\multicolumn{7}{|c|} {\textit{Score only}} \\
\hline

 ST  & \(2.3\%_{\pm 0.5\%}\) & \(2.2\%_{\pm 0.2\%}\) & \(1.92\%_{ \pm 0.6\%}\) & \(1.92\%_{ \pm 0.6\%}\) & \(17.6\%_{ \pm 9.7\%}\) & \(17.6\%_{ \pm 9.7\%}\) \\
\hline
\multicolumn{7}{|c|} {\textit{Score and Uncertainty based}} \\
\hline
% & To-Do & To-Do \\
% Score & 
HR & \(1.2\%_{\pm 1.1\%}\) & \(0.8\%_{\pm 0.7\%}\) & \(0.4\%_{\pm 0.4\%}\) & \(0.4\%_{\pm 0.4\%}\)  & \(11.5\%_{\pm 9.8\%}\) & \(11.5\%_{\pm 9.8\%}\) \\

% HR & \(1.2\%\_{\pm 1.1\%}\) & \(0.8\%\_{\pm 0.7\%}\) & \(0.4\%\_{\pm 0.4\%}\) & \(0.4\%\_{\pm 0.4\%}\)  & \(11.5\%\_{\pm 9.8\%}\) & \(11.5\%\_{\pm 9.8\%}\) \\
%\citep{heuristic2d2022} 
% & To-Do & To-Do \\
% GMT & \(2.4\%\_{\pm 0.5\%}\) & \(2.6\%\_{\pm 0.3\%}\) & \(2.6\%\_{\pm 0.3\%}\) & \(2.6\%\_{\pm 0.3\%}\)  & \(17.8\%\_{\pm 8.7\%}\) & \(20.3\%\_{\pm 6.7\%}\) \\
GMT & \(2.4\%_{\pm 0.5\%}\) & \(2.6\%_{\pm 0.3\%}\) & \(2.6\%_{\pm 0.3\%}\) & \(2.6\%_{\pm 0.3\%}\)  & \(17.8\%_{\pm 8.7\%}\) & \(20.3\%_{\pm 6.7\%}\) \\

% & To-Do & To-Do \\ 
% MIST & \(2.5\%\_{\pm 0.2\%}\) & \(2.7\%\_{\pm 0.3\%}\) & \(2.7\%\_{\pm 0.3\%}\) & \(2.7\%\_{\pm 0.3\%}\)  & \(18.7\%\_{\pm 9.2\%}\) & \(21.6\%\_{\pm 6.7\%}\) \\
MIST & \(2.5\%_{\pm 0.2\%}\) & \(2.7\%_{\pm 0.3\%}\) & \(2.7\%_{\pm 0.3\%}\) & \(2.7\%_{\pm 0.3\%}\)  & \(18.7\%_{\pm 9.2\%}\) & \(21.6\%_{\pm 6.7\%}\) \\

%  MIST & \(2.52\%\_{\pm 0.2\%}\) & \(2.72\%\_{\pm 0.3\%}\) & \(2.72\%\_{\pm 0.3\%}\) & \(2.72\%\_{\pm 0.3\%}\)  & \(18.70\%\_{\pm 9.2\%}\) & \(21.6\%\_{\pm 6.6\%}\) \\
%  & To-Do & To-Do \\ 

% EW-DPMT & - & \(2.7\%\_{\pm 0.3\%}\) & - & \(2.7\%\_{\pm 0.3\%}\)  & - & \(22.3\%\_{\pm 6.7\%}\) \\
EW-DPMT & - & \(2.7\%_{\pm 0.3\%}\) & - & \(2.7\%_{\pm 0.3\%}\)  & - & \(22.3\%_{\pm 6.7\%}\) \\

VW-DPMT & \(2.7\%_{\pm 0.3\%}\) & \(2.7\%_{\pm 0.3\%}\) & \(2.4\%_{\pm 0.3\%}\) & \(2.4\%_{\pm 0.3\%}\)  & \(20.0\%_{\pm 8.7\%}\) & \(22.3\%_{\pm 6.3\%}\) \\

% VWDPMT & \(2.7\%\_{\pm 0.3\%}\) & \(2.7\%\_{\pm 0.3\%}\) & \(2.4\%\_{\pm 0.3\%}\) & \(2.4\%\_{\pm 0.3\%}\)  & \(20.0\%\_{\pm 8.7\%}\) & \(22.3\%\_{\pm 6.3\%}\) \\
%  EW-DPMT & - & 2.7\%_{\pm 0.3\%} & - & 2.7\%_{\pm 0.3\%}  & - & 22.3\%_{ \pm 6.7\%}  \\
%  & To-Do & To-Do \\ 
%  VW-DPMT & 2.7\%_{\pm 0.3\%} & 2.7\%_{\pm 0.3\%} & 2.4\%_{\pm 0.3\%} & 2.4\%_{\pm 0.3\%}  & 20.0\%_{ \pm 8.7\%} & 22.3\%_{ \pm 6.3\%} \\
%  & To-Do & To-Do \\ 
\hline
\end{tabular}
\end{small}
\caption{\footnotesize Recall@PrecisionBound of various decision boundary methods on \texttt{Criteo}, \texttt{Avazu} \& \texttt{E-Com} data. }
%\caption{\footnotesize Performance of different decision boundary algorithms as measured by Recall@PrecisionBound on \texttt{Criteo}, \texttt{Avazu} and \texttt{E-Com} test datasets. }
\label{tab:algo-comparison}
\end{center}
% \end{minipage}\hfill
% \begin{minipage}[b]{0.25\textwidth}
% \centering
% \vspace{-3 mm}
% \includegraphics[width=36mm, height=32mm]{images/extratp/mnist_agg_extratp_ew_fix.pdf}
% \captionof{figure}{\footnotesize 
% \texttt{MNIST-bin} digits where \textsc{EW-DPMT} differs from \textsc{ST} for $70\%$ precision. \textsc{EW-DPMT} includes the top row and excludes the bottom ones. }
% %Qualitative analysis of the output from baseline \textsc{SC} and \textsc{DPMT} algorithms: S
% \label{fig:image}
% \end{minipage}
% \end{center}
%\vskip -0.1in
\vspace{-2mm}
\end{table*}

\begin{comment}
\begin{figure}[!htbp]
    \centering
    \includegraphics[width=0.9\textwidth, height=3cm]{images/extratp/criteo_swapset_pie_charts_redo_c19_rrpt.png}
    \caption{Distribution of subsets of data from \texttt{Criteo} with $\tau=3$ across a key categorical feature (C19): (a) Predicted positive samples as per the 90\% precision decision boundary by EW-DPMT, (b) Bin  included by \textsc{EW-DPMT} but excluded  by \textsc{ST},
    %Incremental True positives flagged by \textsc{EW-DPMT}, 
   (c) Samples from the score bin corresponding to the bin (b) across all uncertainty levels,   (d) Bin excluded by \textsc{EW-DPMT} but included by \textsc{ST},
    and (e) Samples from the score bin corresponding to the bin (d) across all uncertainty levels. 
    Here, V1 refers to C19 with value 1533924, V2 with 1533929, V3 with 1533925.}
    \label{fig:swapset}
\end{figure}

\end{comment}

\begin{figure*}[!htbp]
    \centering
    \hfill
        \subfigure[]{\includegraphics[width=0.31\textwidth, height=3.3cm]{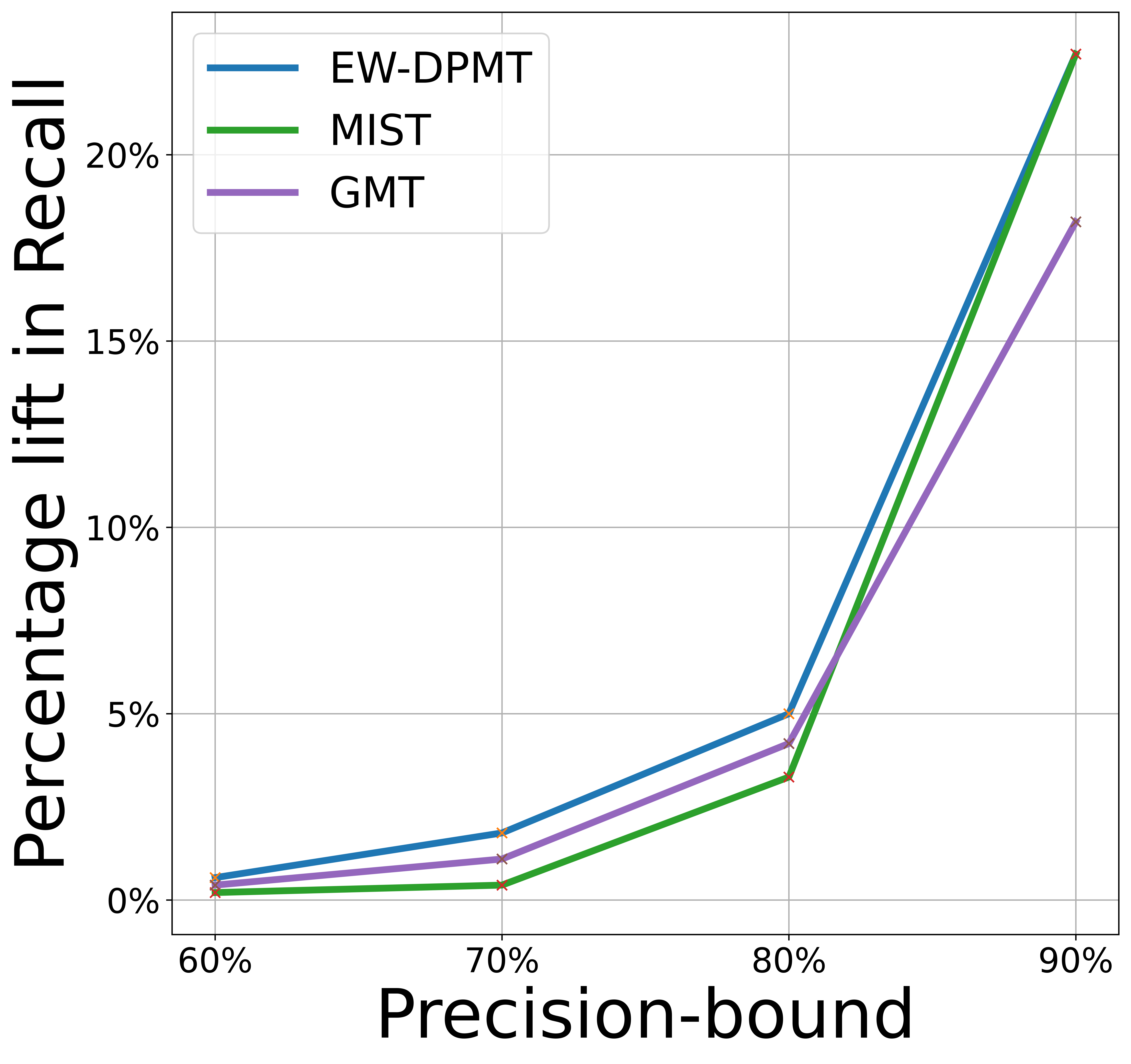}} 
    \hfill
% \subfigure[]{\includegraphics[width=0.24\textwidth]{images/prcurve/mnist_prcurve_17_big_2.pdf}}
    \subfigure[]{\includegraphics[width=0.31\textwidth, height=3.3cm]{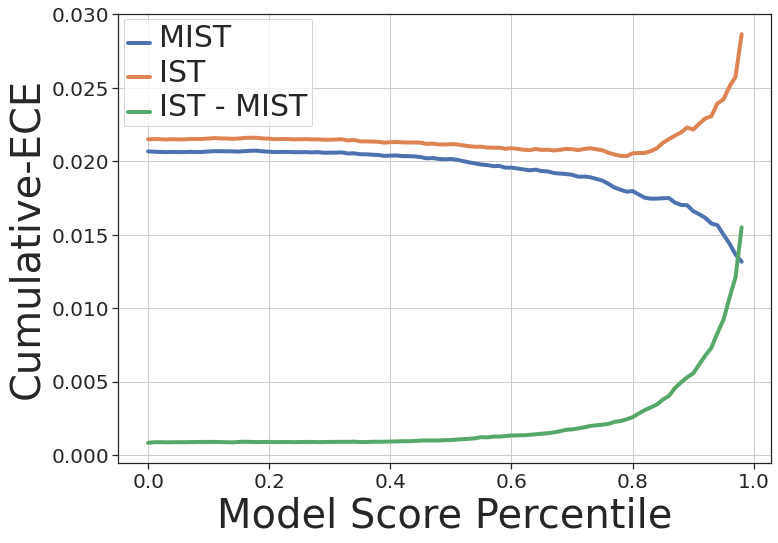}} 
    \hfill
    \vspace{-2mm}
    \caption{\footnotesize (a) Relative gain in recall on uncertainty-based 2-D decision boundary algorithms over the baseline \textsc{ST} on \texttt{Criteo} dataset ($\tau=3$) for different precision bounds. (b) Cumulative Expected Calibration Error(ECE) between MIST and IST baseline  on \texttt{Avazu} dataset ($\tau=5$).}
    \label{fig:ece_multiprec} 
\vspace{-2mm}
\end{figure*}

\noindent{\bf RQ2: Relative Efficacy of Decision boundary Algorithms.}
%\noindent {\bf Recall@Precision Bound.} 
Table \ref{tab:algo-comparison} shows the recall at high precision bounds for various decision boundary algorithms on three large-scale datasets with 500 score and 3 uncertainty bins, averaged over 5 runs with different seeds. Since \texttt{Avazu} and \texttt{E-com} did not have feasible operating points at 90\% precision, we measured recall$@$70\% precision. Across all the datasets, we observe a significant benefit when incorporating uncertainty in the decision boundary selection (paired t-test significance p-values in Table \ref{tab:statsig}). %especially in high score ranges 
%For instance, 
At \@ 90\% precision, \textsc{EW-DPMT} on \texttt{Criteo} is able to achieve a 22\% higher recall (2.7\% vs 2.2\%) over $ST$. Similar behavior is observed on \texttt{Avazu} and \texttt{E-com} datasets where the relative recall lift is 42\% and 26\% respectively. Further, the \texttt{Equi-weight} binning results in more generalizable boundaries with the best performance coming from the DP algorithms (\textsc{EW-DPMT}, \textsc{VW-DPMT}) and the isotonic regression-based MIST. The heuristic baseline \textsc{HR} \citep{heuristic2d2022} performs poorly since it implicitly assumes that positivity rate monotonically increases with uncertainty for a fixed score. 
While  both \textsc{EW-DPMT} and \textsc{MIST} took similar time ($\sim$ 100s) for 500 score bins and 3 uncertainty bins, the run-time of the former increases significantly with increase in the bin count.
 Considering the excessive computation required for \textsc{VW-DPMT}, isotonic regression-based algorithm \textsc{MIST} and \textsc{EW-DPMT} seem to be efficient practical solutions. Results on statistical 
 statistical significance and runtime comparison are in
 Appendix \ref{appendix:statsig} and Appendix \ref{appendix:runtime}. Fig. \ref{fig:ece_multiprec}(a) and Table \ref{tab:recall_with_precision} show the  gain in recall for uncertainty-based 2D-decision boundary algorithms relative to the baseline algorithm  \textsc{ST} highlighting that the increase is larger for high precision range and decreases as the precision level is reduced. Experiments with other uncertainty methods such as MC-Dropout \citep{mcdropout} (see Table \ref{tab:algo-comparison-mcd} )
also point to some but not consistent potential benefits possibly because Posterior networks capture both epistemic and aleatoric uncertainty while MC-Dropout is restricted to the former. 

\begin{figure*}[!htbp]
    \centering
    \hfill
    \subfigure[]{\includegraphics[width=0.29\textwidth, height=3.5cm]{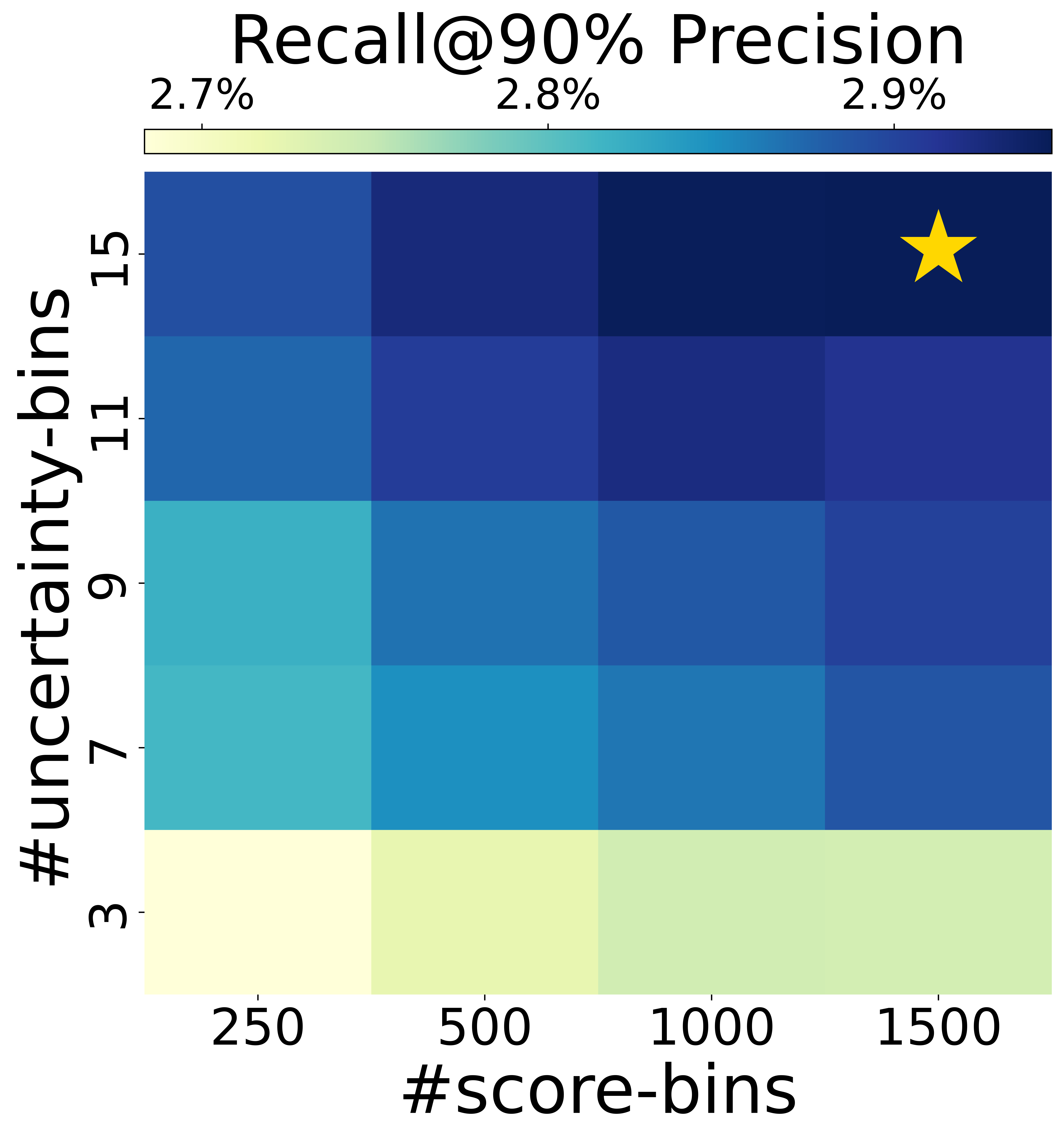}} 
    \hfill
    \subfigure[]{\includegraphics[width=0.29\textwidth, height=3.5cm]{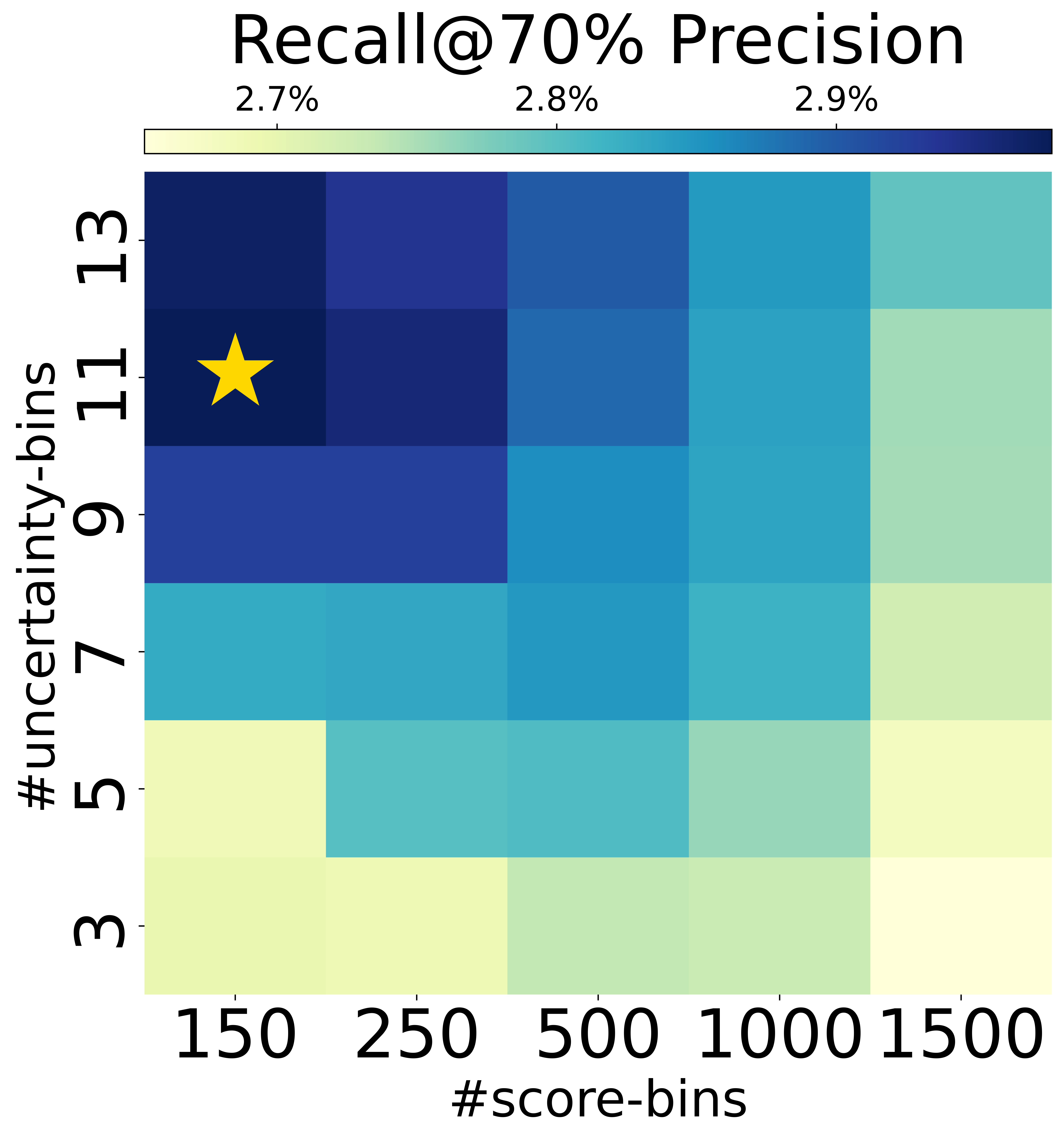}} 
    \hfill
    % \subfigure[]{\includegraphics[width=0.24\textwidth]{images/prcurve/mnist_prcurve_17_big_2.pdf}}
    \subfigure[]{\includegraphics[width=0.29\textwidth, height=3.4cm]{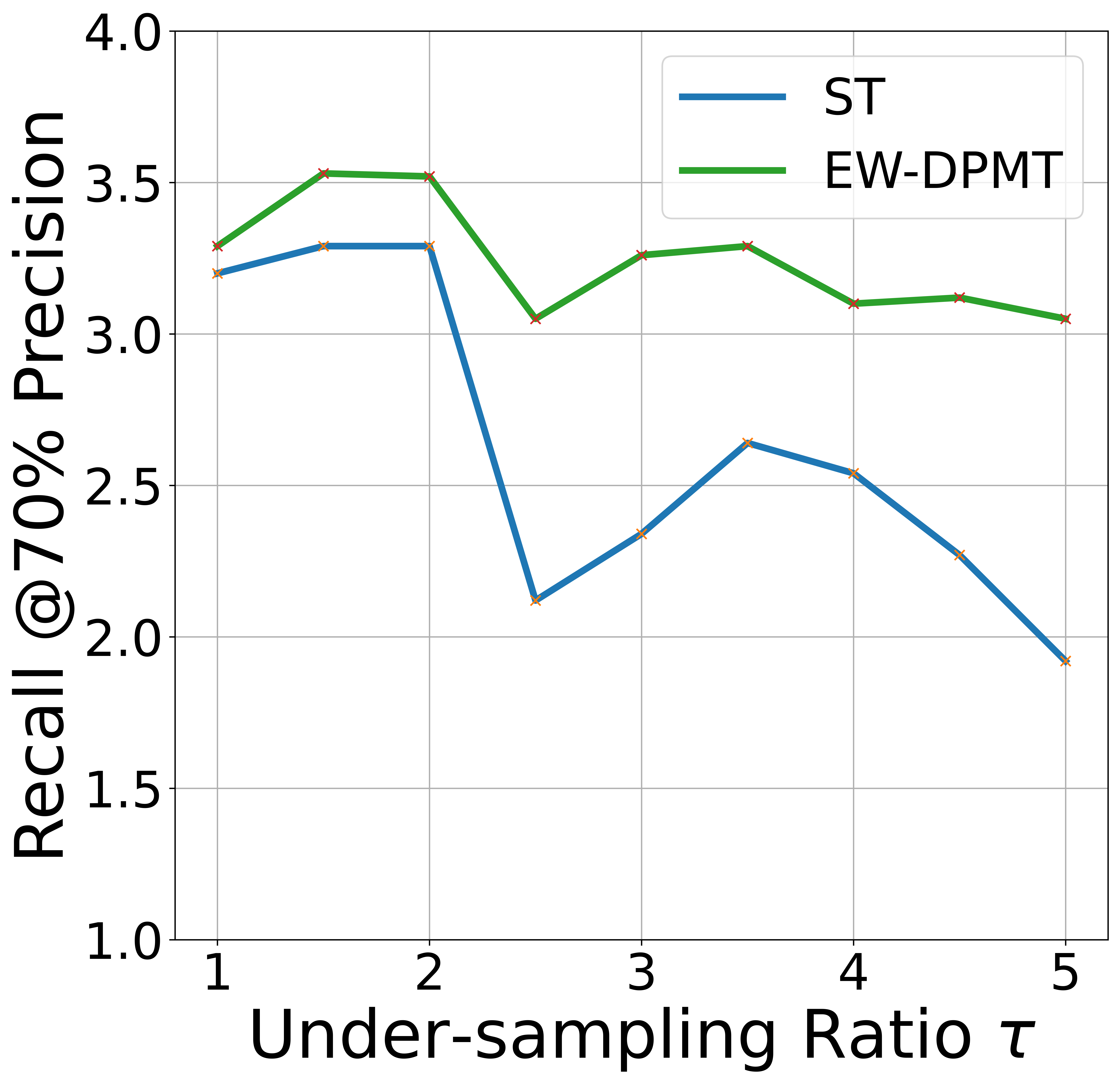}}
    \hfill
    \vspace{-2mm}
    \caption{\footnotesize Impact of \# bins along uncertainty and score for \textsc{EW-DPMT} on (a) \texttt{Criteo} ($\tau=3$, Recall@90\% Precision) and (b) \texttt{Avazu} ($\tau=5$, Recall@70\%Precision). (c) Impact of undersampling level ($\tau$) during training on Recall@70\%Precision for \textsc{ST} and \textsc{EW-DPMT} on \texttt{Avazu}.}
    \label{fig:RQ3}
\vspace{-1mm}
\end{figure*}

% TMPCOMMENTING
% Fig \ref{fig:ECE} in Appendix \ref{appendix:ece} demonstrates that the difference between ECE for \textsc{MIST} and \textsc{IST} increases as we move towards higher score range.
% We discuss statistical significance of results in Appendix \ref{appendix:statsig} and algorithm runtimes in Appendix \ref{appendix:runtime}. 

\noindent{\bf RQ3: Dependence on choice of bins and undersampling ratio.}

\noindent {\bf Binning configuration.} Fig. \ref{fig:RQ3}(a) and \ref{fig:RQ3}(b) show how performance (Recall@PrecisionBound) of \textsc{EW-DPMT}  %algorithm 
varies with the number of uncertainty and score bins for \texttt{Criteo} and \texttt{Avazu} datasets. We observe 
%that there exists
a dataset dependent sweet-spot (marked by star) for  the choice of bins. 
%that is dataset dependent. 
%a  sweet-spot (marked by star) for  the choice of bins that is dataset dependent.  
Too many bins 
%Over-binning (i.e., too many bins) 
can lead to overfitting of the decision boundary on the hold-out set that does not generalize well to test setting, while under-binning leads to low recall improvements on both hold-out and test sets. 
%We explore different ranges of bins for \texttt{Criteo} and \texttt{Avazu} due to varying dataset sizes. %since the former has fewer samples.

\noindent{\bf Undersampling Ratio.} Fig. \ref{fig:RQ3}(c) captures the Recall@70\% Precision performance of \textsc{EW-DPMT} and \textsc{ST} for different levels of undersampling ($\tau$) of the negative class on the \texttt{Avazu} dataset averaged over 5 seeds. For both the algorithms, we observe an improvement 
%when there is downsampling of negatives benefits 
in recall performance initially (till $\tau=2.5$) which disappears for higher levels of downsampling in accordance with prior studies~\citep{downsampling}. We observe that \textsc{EW-DPMT} consistently improves the Recall@70\% precision over \textsc{ST} with more pronounced  downsampling (i.e., higher values of $\tau$). 
%The improvements increase with higher values of $\tau$.

%For each use-case, we choose the precision level based on the Precision-Recall curves to achieve reasonable recall at the level. Thus we chose 90\% for \texttt{Criteo}, 70\% for \texttt{Avazu} and \texttt{E-Com} datasets. 
\noindent{\bf RQ4: Impact of leveraging uncertainty for probability calibration.}
To investigate  the potential benefits of incorporating uncertainty in improving probability calibration, we compared the probabilities output from \textsc{MIST} algorithm with those  from a vanilla isotonic regression (\textsc{IST}) baseline on Expected Calibration Error (ECE) for every score-bin, averaged across different uncertainty levels. 
%Fig. \ref{fig:ece_multiprec} (a) and
Fig. \ref{fig:ECE} (b) demonstrates that the difference between ECE for \textsc{MIST} and \textsc{IST} increases as we move towards higher score range. Thus, the benefit of leveraging uncertainty estimates in calibration is more pronounced in high score range (i.e. at high precision levels). More details in Appendix \ref{appendix:ece}.

\section{Conclusion}
\label{sec:conclusion}

%O:the problem we studied
%O:contributions and takeaways
%O: future directions
Leveraging uncertainty estimates for ML-driven decision-making is an important area of research. In this paper, we investigated potential benefits of utilizing  uncertainty along with model score for binary classification. We provided theoretical analysis that points to the discriminating ability of  uncertainty and formulated a novel 2-D decision boundary estimation problem based on score and uncertainty that turns out to be NP-hard. We also proposed practical algorithmic solutions based on dynamic programming and isotonic regression. Empirical evaluation on real-world datasets point to the efficacy 
of utilizing uncertainty  in improving classification performance. Future directions of exploration include (a) designing efficient algorithms for joint optimization of binning configuration and boundary detection, (b) utilizing uncertainty for improving  ranking performance and explore-exploit strategies in applications such as recommendations where the relative ranking matters and addressing data bias is critical,  and (c) extensions to regression and multi-class classification settings.

\bibliography{iclr2024_conference}
\bibliographystyle{iclr2024_conference}
\newpage
\appendix
\section{Reproducibility Statement}
\label{appendix:repro_state}
To ensure the reproducibility of our experiments, we provide details of hyperparameters used for training posterior network model with details of model (backbone used and flow parameters) in Sec. \ref{subsec:exp_setup}. All models were trained on NVIDIA 16GB V100 GPU. We provide the pseudo code of binning and all algorithms implemented in Sec. \ref{sec:algo} and Appendix \ref{appendix:algorithm} with details of bin-configuration in Sec \ref{subsec:results}. All binning and decision boundary related operations were performed on 4-core machine using Intel Xeon processor 2.3 GHz (Broadwell E5-2686 v4) running Linux. Moreover, we will publicly open-source our code later after we cleanup our code package and add proper documentation for it. 
\section{Ethics Statement}
Our work is in accordance with the recommended ethical guidelines. 
Our experiments are performed on three datasets, two of which are well-known click prediction Datasets (\texttt{Criteo}, \texttt{Avazu}) datasets in public domain. The third one is a proprietary dataset related to customer actions but collected with 
explicit consent of the customers while adhering to strict customer 
data confidentiality and security guidelines. The data we use is anonmyized by one-way hashing. Our proposed methods are targeted towards classification performance for any generic classifier and carry the risks common to all AI techniques.
\section{Additional Experimental Results}

\subsection{Benefits of 2D-Decision Boundary Estimation}
\label{appendix:incremental_pos}
\begin{figure}[!htbp]
    \centering
    \includegraphics[width=0.98\textwidth, height=4.5cm]{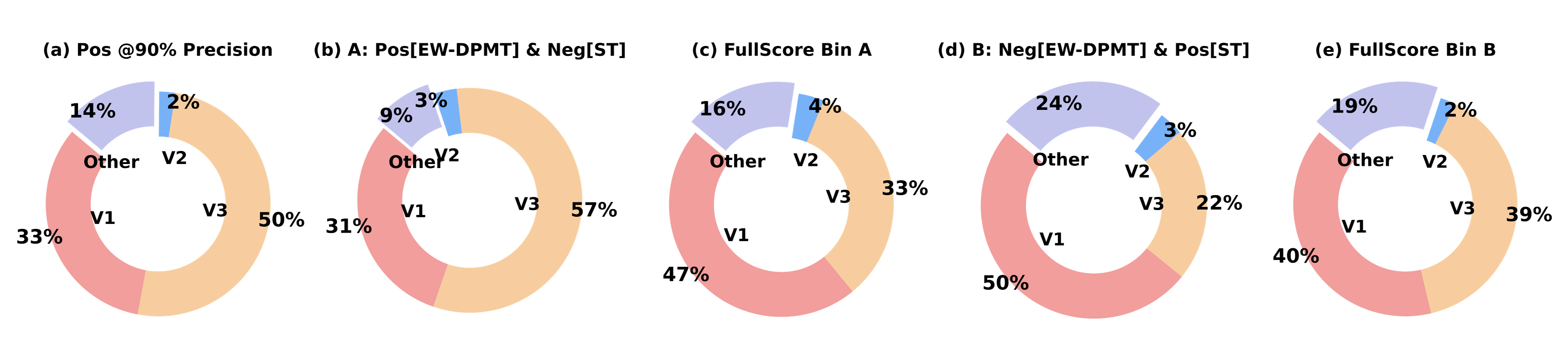}
    \caption{Distribution of subsets of data from \texttt{Criteo} with $\tau=3$ across a key categorical feature (C19): (a) All positive samples as per the 90\% precision decision boundary by EW-DPMT, (b) Bin A included by \textsc{EW-DPMT} in the positive region but excluded  by \textsc{ST},
    %Incremental True positives flagged by \textsc{EW-DPMT}, 
    (c) Score bin corresponding to the bin A across all uncertainty levels, 
    %Samples for  score-range where uncertainty-based bins were excluded in \textsc{EW-DPMT} but included by \textsc{ST}
    (d) Bin B excluded by \textsc{EW-DPMT} in the positive region but included by \textsc{ST},
    and (e) Score bin corresponding to the bin B across all uncertainty levels. 
 %Samples from the score-range where uncertainty-based bins were included by \textsc{EW-DPMT} but excluded by \textsc{ST}. 
    Here, V1 refers to C19 with value 1533924, V2 with 1533929, V3 with 1533925.}
    \label{fig:swapset}
\end{figure}

To anecdotally validate the benefits of 2D decision boundary estimation, we run  
the algorithms \textsc{EW-DPMT} and \textsc{ST}  on \texttt{Criteo} dataset and examine bins where the algorithms   differ on the  decision boundary  for 90\% precision. As mentioned earlier, the bin (bin $A$) with
 $[(s(\x), u(\x))=(0.984, 0)]$ and positivity rate $0.91$  is included in the positive region by \textsc{EW-DPMT} but excluded by  \textsc{ST} while the reverse is true for the bin (bin $B$) with $[(s(\x), u(\x))=(0.996, 0.667)]$ and positivity rate $0.87$. Note that $((s(\x), u(\x))$ are the score and uncertainty percentiles and not the actual values. We further characterise these bins using informative  categorical features. Fig. \ref{fig:swapset} depicts pie-charts of the feature distribution of one of these features "C19" for both these bins as well as the corresponding score bins across all uncertainty levels and  the entire positive region as identified by \textsc{EW-DPMT}.  From the plots, we observe that the distribution of C19 for the positive region of 
\textsc{EW-DPMT} (Fig. \ref{fig:swapset} (a)) is  similar  to that of the  bin A (Fig. \ref{fig:swapset} (b)) which 
is labeled positive by \textsc{EW-DPMT} and negative by \textsc{ST} and different from that of bin B (Fig. \ref{fig:swapset} (c)) that is labeled negative  by  \textsc{EW-DPMT} but positive by \textsc{ST}  in terms of feature value V1 being more prevalent in the latter. 
We also observe that  bins A and B diverge from the corresponding entire score bins across uncertainty-levels, i.e., Fig. \ref{fig:swapset} (c) and Fig. \ref{fig:swapset} (e) respectively.  This variation of both feature distribution and  positivity with uncertainty for the same score range highlights the need for flexible 2D decision boundary estimation beyond vanilla thresholding based on score alone.

% \hfill

\subsection{Recall improvement at different precision levels}
\label{appendix:recall_with_precision}
To identify the precision regime where the 2D-decision boundary algorithms are beneficial, we  measure the recall from the various algorithms for different precision bounds. Table \ref{tab:recall_with_precision} shows the results on \textsc{Criteo} dataset ($\tau=3$) highlighting that the relative improvement by leveraging uncertainty estimation in decision boundary estimation increases with precision bound. This empirically ties to the observation that the separation between different uncertainty levels is more prominent for higher score range, as this separation is used by 2D-decision boundary algorithms for improving recall.

%To understand how recall improvement varies with the choice of precision bound, we measured the recall from the decision boundary algorithms for different choices of precision. For \textsc{Criteo} dataset ($\tau=3$) setting, Table \ref{tab:recall_with_precision} highlights that the relative improvement in recall by leveraging uncertainty estimation in decision boundary estimation is increases with Precision at which it was evaluated. This empirically ties to the observation that the separation between different uncertainty levels is more prominent for higher score range, as this separation is used by 2D-decision boundary algorithms for improving recall.
\begin{table*}[!htbp]%[!htbp]
\begin{center}
\begin{small}

\begin{tabular}{|c||c|c|c|c|}
\hline
\footnotesize
% \hline
\textbf{Algorithm} & \textbf{Recall@}
 & \textbf{Recall@}  & \textbf{Recall@}  & \textbf{Recall@}
\\
 & \textbf{60\%Precision} & 
\textbf{70\%Precision} & \textbf{80\%Precision} & \textbf{90\%Precision} \\
% \multicolumn{7}{|c|} {\textit{Score only}} \\
\hline
\multicolumn{5}{|c|} {\textit{Score only}}\\
\hline
 ST  & \(46.6\%_{\pm 0.9\%}\)  & \(27.7\%_{\pm 1.1\%}\)  & \(12.0\%_{\pm 0.8\%}\) & \(2.2\%_{\pm 0.2\%}\) \\
\hline
\multicolumn{5}{|c|} {\textit{Score and Uncertainty based}} \\
\hline
GMT  & \(46.8\%_{\pm 0.9\%}\) (\textbf{+0.4\%})  & \(28.0\%_{\pm 1.0\%}\) (\textbf{+1.0\%}) & \(12.5\%_{\pm 0.8\%}\) (\textbf{+3.5\%}) & \(2.6\%_{\pm 0.3\%}\)  (\textbf{+18.2\%})\\
MIST  & \(46.7\%_{\pm 0.9\%}\) (\textbf{+0.1\%})  & \(27.8\%_{\pm 1.0\%}\) (\textbf{+0.4\%})  & \(12.4\%_{\pm 0.6\%}\) (\textbf{+2.7\%}) & \(2.7\%_{\pm 0.3\%}\) (\textbf{+22.7\%}) \\
EW-DPMT & \(46.9\%_{\pm 0.8\%}\) (\textbf{+0.6\%})  & \(28.2\%_{\pm 1.0\%}\) (\textbf{+1.6\%}) & \(12.6\%_{\pm 0.8\%}\) (\textbf{+4.8\%}) & \(2.7\%_{\pm 0.3\%}\) (\textbf{+22.7\%}) \\
\hline
\end{tabular}
\end{small}
\caption{\footnotesize Recall@ different precision levels for \texttt{Criteo} dataset ($\tau=3$) for various decision boundary algorithms along with relative gains for each uncertainty level (in brackets)  relative to the \textsc{ST} algorithm.}
\label{tab:recall_with_precision}
\end{center}
% \end{minipage}\hfill
% \begin{minipage}[b]{0.25\textwidth}
% \centering
% \vspace{-3 mm}
% \includegraphics[width=36mm, height=32mm]{images/extratp/mnist_agg_extratp_ew_fix.pdf}
% \captionof{figure}{\footnotesize 
% \texttt{MNIST-bin} digits where \textsc{EW-DPMT} differs from \textsc{ST} for $70\%$ precision. \textsc{EW-DPMT} includes the top row and excludes the bottom ones. }
% %Qualitative analysis of the output from baseline \textsc{SC} and \textsc{DPMT} algorithms: S
% \label{fig:image}
% \end{minipage}
% \end{center}
%\vskip -0.1in
% \vspace{-3mm}
\end{table*}

% \begin{figure}[!htbp]
%     \centering
%     \includegraphics[width=0.8\textwidth, height=5.2cm]{images/prcurve/multiprec_criteo_tau3.png}
%     \caption{Relative improvement (in \%age) of different uncertainty incorporating decision boundary algorithms over the single score based threshold \textsc{ST} at different precision levels for \textsc{Criteo} dataset $\tau=3$}
%     \label{fig:swapset}
% \end{figure}

\subsection{Statistical Significance of various algorithms vs ST}
\label{appendix:statsig}
Table \ref{tab:statsig} captures the significance levels in the form of p-values on paired t-test (one-sided) comparing the different algorithms against the single-threshold (\textsc{ST}). It is evident that  algorithms that leverage both score and uncertainty such as \textsc{EW-DPMT}, \textsc{MIST}, \textsc{VW-DPMT} and \textsc{GMT}  significantly outperform \textsc{ST}, improving recall at fixed precision for all datasets.
\begin{table*}[!htbp]%[!htbp]
% \vspace{-2mm}
% \begin{minipage}[b]{0.6\textwidth}
\begin{center}
\begin{footnotesize}
\begin{tabular}{|l||c|c||c|c||c|c|}
\hline

  & \multicolumn{2}{|c||}{\texttt{Criteo, 90\% Precision}} & \multicolumn{2}{|c||}{\texttt{Avazu, 70\% Precision}} & \multicolumn{2}{|c|}{\texttt{E-Com, 70\% Precision}} \\
%  & \multicolumn{2}{|c|}{CIFAR10-bin} \\
 \hline
 \hline
 & \multicolumn{2}{|c||}{\small{$\tau$=3, Pos:Neg = 1:3}} & \multicolumn{2}{|c||}{\small{$\tau$=5, Pos:Neg = 1:5}}& \multicolumn{2}{|c|}{\small{$\tau$=5, Pos:Neg = 1:24}} \\
\hline

\hline
 \textbf{Algorithm} & \textbf{Equi-Span} & \textbf{Equi-weight} & \textbf{Equi-Span} & \textbf{Equi-weight} & \textbf{Equi-Span} & \textbf{Equi-weight} \\
% & Equi-Span & Equi-weight \\
\hline
% \multicolumn{7}{|c|} {\textit{Score only}} \\
% \hline
%  ST  & 2.3_{\pm 0.5} & 2.2\%_{\pm 0.2\%} & 1.92\%_{ \pm 0.6\%} & 1.92\%_{ \pm 0.6\%} & x & x \\
% \hline
\multicolumn{7}{|c|} {\textit{Score and Uncertainty based}} \\
\hline
% & To-Do & To-Do \\
% Score & 
ST vs. HR & 0.99 & 0.98 & 0.99 & 0.99  &  &  \\
%\citep{heuristic2d2022} 
% & To-Do & To-Do \\
ST vs. GMT &  & 0.08 & 0.03 & 0.03 & 0.42 & 0.07 \\
% & To-Do & To-Do \\ 
ST vs. MIST & 0.05 & 0.03 & 0.02 & 0.02 & 0.1 & 0.03 \\
%  MIST & 2.52\%_{\pm 0.2\%} & 2.72\%_{\pm 0.3\%} & 2.72\%_{\pm 0.3\%} & 2.72\%_{\pm 0.3\%}  & 18.70\%_{\pm 9.2\%} & 21.6\%_{\pm 6.6\%} \\
%  & To-Do & To-Do \\ 
ST vs. EW-DPMT & - & 0.03 & - & 0.02  & - & 0.01 \\
%  & To-Do & To-Do \\ 
ST vs. VW-DPMT & 0.03 & 0.03 & 0.02 & 0.02  & 0.02 & 0.01 \\
%  & To-Do & To-Do \\ 
\hline
\end{tabular}
\end{footnotesize}
\caption{\footnotesize Significance level (p-values) of Paired t-test on Recall@PrecisionBound of different decision boundary algos on \texttt{Criteo}, \texttt{Avazu} and \texttt{E-Com} test datasets with \#bins chosen same as per Table \ref{tab:algo-comparison}.}
\label{tab:statsig}
\end{center}
% \end{minipage}\hfill
% \begin{minipage}[b]{0.25\textwidth}
% \centering
% \vspace{-3 mm}
% \includegraphics[width=36mm, height=32mm]{images/extratp/mnist_agg_extratp_ew_fix.pdf}
% \captionof{figure}{\footnotesize 
% \texttt{MNIST-bin} digits where \textsc{EW-DPMT} differs from \textsc{ST} for $70\%$ precision. \textsc{EW-DPMT} includes the top row and excludes the bottom ones. }
% %Qualitative analysis of the output from baseline \textsc{SC} and \textsc{DPMT} algorithms: S
% \label{fig:image}
% \end{minipage}
% \end{center}
%\vskip -0.1in
% \vspace{-3mm}
\end{table*}

% \begin{comment}

\subsection{Runtime of various algorithms}
\label{appendix:runtime}
Table \ref{tab:runtime} shows the run-times (in seconds) for the best performing algorithms: MIST (Multi Isotonic regression Single score Threshold) and EW-DPMT (Equi-Weight Dynamic Programming based Multi threshold on \texttt{Criteo} ($\tau=3$) dataset for different bin sizes with 64-core machine using Intel Xeon processor 2.3 GHz (Broadwell E5-2686 v4) running Linux. The runtimes are averaged over 5 experiment-seeds for each setting. 
% As mentioned in the paper, the calibration split of this dataset has XX instances. 
The run-times do not include the binning time since  this is the same for all the algorithms for a given  binning configuration. It only includes the time taken to fit the decision boundary algorithm and obtain Recall@PrecisionBound.

%Table \ref{tab:runtime} shows the run-times (in seconds) for the best performing algorithms: MIST (Multi Isotonic regression Single score Threshold) and EW-DPMT (Equi-Weight Dynamic Programming based Multi threshold on \texttt{Criteo} ($\tau=3$) dataset for different bin sizes with 64-core machine using Intel Xeon processor 2.3 GHz (Broadwell E5-2686 v4) running Linux. As mentioned in the paper, the calibration split of this dataset has  instances. The run-times do not include the binning time as for each score, uncertainty binning, the time-taken for each algorithm is same. Hence the run-time only includes the time taken to fit the decision boundary algorithm and obtain Recall@PrecisionBound. The run-times are averaged over 5 experiment-seeds for each setting. 

\begin{table}[!ht]
\caption{Wallclock runtime (in seconds) of various algorithms on \texttt{Criteo} Dataset ($\tau=3$).} 
    \centering
    \label{tab:runtime}
\begin{tabular}{|c||c|c|c|c|c|c|}
% \hline & \multicolumn{6}{|c|}{Score-bins} \\
\hline
 Score-bins & \multicolumn{2}{|c|}{100} & \multicolumn{2}{|c|}{500} & \multicolumn{2}{|c|}{1000} \\
\hline
Uncertainty-bins & EW-DPMT & MIST & EW-DPMT & MIST & EW-DPMT & MIST \\
\hline
3  & 8       & 92   & 98      & 99   & 530     & 80 \\
\hline
7  & 42      & 154  & 830     & 146  & 3434    & 99 \\
\hline
11 & 106     & 233  & 1976    & 162  & 11335   & 99 \\
\hline
\end{tabular}
\end{table}

From the theoretical analysis, we expect the runtime of decision boundary estimation for MIST to be $O(KL \log(KL) )$. However, in practice there is a strong dependence only on $K$ i.e., the number of uncertainty bins since we invoke an optimized implementation of isotonic regression $K$ times. Furthermore, we perform the isotonic regression over the samples directly instead of the aggregates over the
$L$ score bins  which reduces the dependence on $L$.
%effectively making $L=N/K$ .
The final sorting that contributes to the $KL\log(KL)$ term is also optimized and does not dominate the run-time. For \textsc{EW-DPMT}, we expect a runtime complexity of $O(K^2L^2)$, i.e., quadratic in the number of bins. From decision-boundary algorithm fitting perspective, the observed run-times show faster than linear yet sub-quadratic growth due to fixed costs and python optimizations. Overall, \textsc{MIST} performs at par with \textsc{EW-DPMT} on the decision quality but takes considerably less time. 

%From the theoretical analysis, we expect the runtime of decision boundary estimation for MIST to be $O(KL \log(KL) )$. However, in practice there is a strong dependence only on $K$ i.e., the number of uncertainty bins since we invoke an optimized implementation of isotonic regression $K$ times. The final sorting that contributes to the $KL\log(KL)$ term is also optimized and does not dominate the run-time. For EW-DPMT, we expect a runtime complexity of $O(K^2L^2)$, i.e., quadratic in the number of bins. Only from decision-boundary algorithm fitting perspective, the observed run-times show faster than linear yet sub-quadratic growth in run-times due to fixed costs, python optimizations etc. Overall, MIST which performs at par with EW-DPMT but takes considerably less time. 
% However, the observed run-times are sub-quadratic due to the fixed costs but grow faster compared to MIST, which makes the latter a preferable option.  
%   \end{comment}

\subsection{Results on MC-Dropout}
To understand the impact of choice of uncertainty estimation method, we report experiments on MC-Dropout \citep{mcdropout} algorithm in Table \ref{tab:algo-comparison}. MC-Dropout estimates epistemic uncertainty of a model by evaluating the variance in output from multiple forward passes of the model for every input sample. Resuts in Table \ref{tab:algo-comparison-mcd} are from models trained for each dataset without any normalizing flow. While we observe substantial relative improvement when the recall is already low as in the case of \texttt{Avazu}, the magnitude of improvement is much smaller than in the case of Posterior Network  possibly because the MC-Dropout uncertainty estimation does not account for aleatoric uncertainty.

\begin{table*}[!htbp]%[!htbp]
% \vspace{-2mm}
% \begin{minipage}[b]{0.6\textwidth}
\begin{center}
\begin{small}
\begin{tabular}{|c||c||c||c|}
\hline
  & \multicolumn{1}{|c||}{\texttt{Criteo}, 90\% Precision} & \multicolumn{1}{|c||}{\texttt{Avazu}, 70\% Precision} & \multicolumn{1}{|c|}{\texttt{E-Com}, 70\% Precision} \\
%  & \multicolumn{2}{|c|}{CIFAR10-bin} \\
 \hline
 \hline
 & \multicolumn{1}{|c||}{\small{$\tau$=3, Pos:Neg = 1:3}} & \multicolumn{1}{|c||}{\small{$\tau$=5, Pos:Neg = 1:5}}& \multicolumn{1}{|c|}{\small{$\tau$=5, Pos:Neg = 1:24}} \\
\hline

\hline
 \textbf{Algorithm} & 
%  \textbf{Equi-Span} & 
 \textbf{Equi-weight} & 
%  \textbf{Equi-Span} & 
 \textbf{Equi-weight} & 
%  \textbf{Equi-Span} & 
 \textbf{Equi-weight} \\
% & Equi-Span & Equi-weight \\
\hline
\multicolumn{4}{|c|} {\textit{Score only}} \\
\hline
 ST  & \(4.2\%_{\pm 0.1\%}\) & \(1.28\%_{ \pm 0.1\%}\) & \(22.8\%_{ \pm 4.7\%}\) \\
\hline
\multicolumn{4}{|c|} {\textit{Score and Uncertainty based}} \\
\hline
% & To-Do & To-Do \\
% Score & 
% HR & \(1.2\%_{\pm 1.1\%}\) & \(0.8\%_{\pm 0.7\%}\) & \(0.4\%_{\pm 0.4\%}\) & \(0.4\%_{\pm 0.4\%}\)  & \(11.5\%_{\pm 9.8\%}\) & \(11.5\%_{\pm 9.8\%}\) \\

% HR & \(1.2\%\_{\pm 1.1\%}\) & \(0.8\%\_{\pm 0.7\%}\) & \(0.4\%\_{\pm 0.4\%}\) & \(0.4\%\_{\pm 0.4\%}\)  & \(11.5\%\_{\pm 9.8\%}\) & \(11.5\%\_{\pm 9.8\%}\) \\
%\citep{heuristic2d2022} 
% & To-Do & To-Do \\
% GMT & \(2.4\%\_{\pm 0.5\%}\) & \(2.6\%\_{\pm 0.3\%}\) & \(2.6\%\_{\pm 0.3\%}\) & \(2.6\%\_{\pm 0.3\%}\)  & \(17.8\%\_{\pm 8.7\%}\) & \(20.3\%\_{\pm 6.7\%}\) \\
% GMT & \(2.4\%_{\pm 0.5\%}\) & \(2.6\%_{\pm 0.3\%}\) & \(2.6\%_{\pm 0.3\%}\) & \(2.6\%_{\pm 0.3\%}\)  & \(17.8\%_{\pm 8.7\%}\) & \(20.3\%_{\pm 6.7\%}\) \\

% & To-Do & To-Do \\ 
% MIST & \(2.5\%\_{\pm 0.2\%}\) & \(2.7\%\_{\pm 0.3\%}\) & \(2.7\%\_{\pm 0.3\%}\) & \(2.7\%\_{\pm 0.3\%}\)  & \(18.7\%\_{\pm 9.2\%}\) & \(21.6\%\_{\pm 6.7\%}\) \\
MIST & \(4.3\%_{\pm 0.2\%}\)(\textbf{+2.4\%})  &  \(1.8\%_{\pm 0.2\%}\) (\textbf{+40\%})& \(23.1\%_{\pm 3.6\%}\)(\textbf{+1.3\%}) \\

%  MIST & \(2.52\%\_{\pm 0.2\%}\) & \(2.72\%\_{\pm 0.3\%}\) & \(2.72\%\_{\pm 0.3\%}\) & \(2.72\%\_{\pm 0.3\%}\)  & \(18.70\%\_{\pm 9.2\%}\) & \(21.6\%\_{\pm 6.6\%}\) \\
%  & To-Do & To-Do \\ 

% EW-DPMT & - & \(2.7\%\_{\pm 0.3\%}\) & - & \(2.7\%\_{\pm 0.3\%}\)  & - & \(22.3\%\_{\pm 6.7\%}\) \\
EW-DPMT & \(4.3\%_{\pm 0.1\%}\)(\textbf{+2.4\%}) & \(1.9\%_{\pm 0.2\%}\)(\textbf{+48\%})  &  \(23.7\%_{\pm 3.8\%}\)(\textbf{+3.9\%}) \\

% VW-DPMT & \(2.7\%_{\pm 0.3\%}\) & \(2.7\%_{\pm 0.3\%}\) & \(2.4\%_{\pm 0.3\%}\) & \(2.4\%_{\pm 0.3\%}\)  & \(20.0\%_{\pm 8.7\%}\) & \(22.3\%_{\pm 6.3\%}\) \\

% VWDPMT & \(2.7\%\_{\pm 0.3\%}\) & \(2.7\%\_{\pm 0.3\%}\) & \(2.4\%\_{\pm 0.3\%}\) & \(2.4\%\_{\pm 0.3\%}\)  & \(20.0\%\_{\pm 8.7\%}\) & \(22.3\%\_{\pm 6.3\%}\) \\
%  EW-DPMT & - & 2.7\%_{\pm 0.3\%} & - & 2.7\%_{\pm 0.3\%}  & - & 22.3\%_{ \pm 6.7\%}  \\
%  & To-Do & To-Do \\ 
%  VW-DPMT & 2.7\%_{\pm 0.3\%} & 2.7\%_{\pm 0.3\%} & 2.4\%_{\pm 0.3\%} & 2.4\%_{\pm 0.3\%}  & 20.0\%_{ \pm 8.7\%} & 22.3\%_{ \pm 6.3\%} \\
%  & To-Do & To-Do \\ 
\hline
\end{tabular}
\end{small}
\caption{\footnotesize Performance of different decision boundary algorithms as measured Recall@PrecisionBound on \texttt{Criteo}, \texttt{Avazu} and \texttt{E-Com} test datasets with MC-Dropout as uncertainty estimation method. }
\label{tab:algo-comparison-mcd}
\end{center}
% \end{minipage}\hfill
% \begin{minipage}[b]{0.25\textwidth}
% \centering
% \vspace{-3 mm}
% \includegraphics[width=36mm, height=32mm]{images/extratp/mnist_agg_extratp_ew_fix.pdf}
% \captionof{figure}{\footnotesize 
% \texttt{MNIST-bin} digits where \textsc{EW-DPMT} differs from \textsc{ST} for $70\%$ precision. \textsc{EW-DPMT} includes the top row and excludes the bottom ones. }
% %Qualitative analysis of the output from baseline \textsc{SC} and \textsc{DPMT} algorithms: S
% \label{fig:image}
% \end{minipage}
% \end{center}
%\vskip -0.1in
\vspace{-2mm}
\end{table*}

\subsection{Impact of using uncertainty estimation on Calibration Error}
\label{appendix:ece}
\begin{figure*}[!htbp]
    \centering
    \subfigure[]{\includegraphics[width=0.24\textwidth, height=2.8cm]{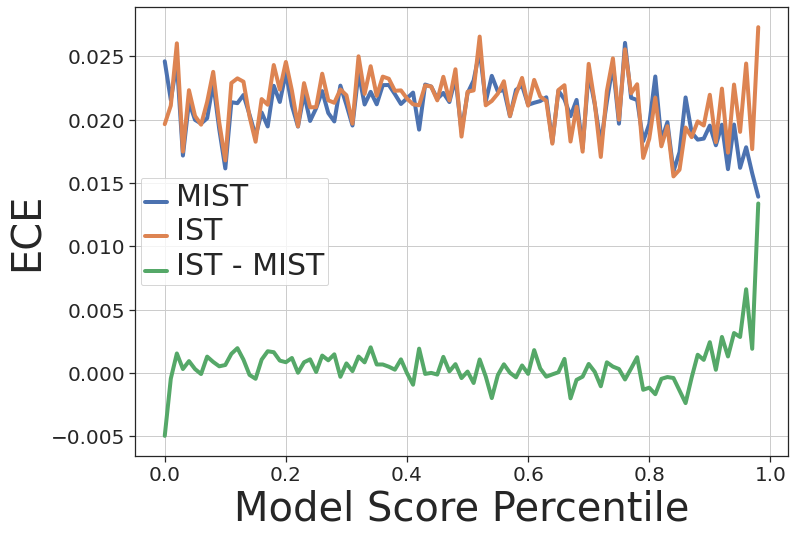}} 
    \subfigure[]{\includegraphics[width=0.24\textwidth, height=2.8cm]{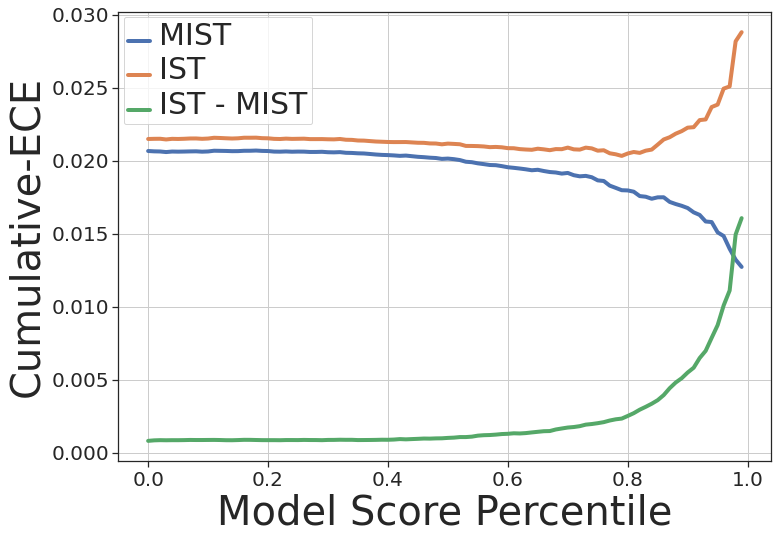}} 
    % \subfigure[]{\includegraphics[width=0.24\textwidth]{images/prcurve/mnist_prcurve_17_big_2.pdf}}
    \subfigure[]{\includegraphics[width=0.24\textwidth, height=2.8cm]{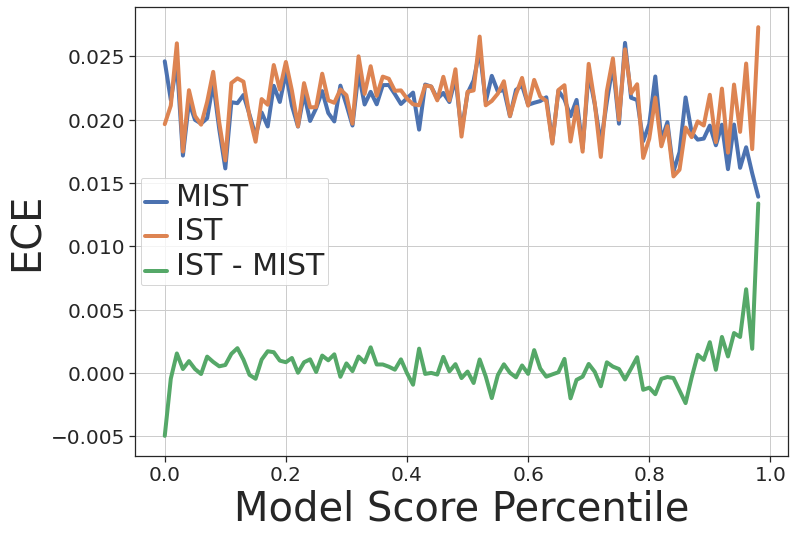}}
    \subfigure[]{\includegraphics[width=0.24\textwidth, height=2.8cm]{images/calibration_error/cumulative_ece_avazu_99_rd2.png}}
    \caption{\footnotesize Impact of leveraging uncertainty in calibration by comparing MIST vs IST (a) $ECE@i$ for \texttt{Criteo} ($\tau=3$) (b) Cumulative-$ECE@j$ for \texttt{Criteo} ($\tau=3$) (c) $ECE@i$ for \texttt{Avazu} ($\tau=5$) (d) Cumulative-$ECE@j$ for \texttt{Avazu} ($\tau=5$).}
    \label{fig:ECE}
\vspace{-1.5mm}
\end{figure*}

For applications such as advertising, it is desirable to have well-calibrated probabilities and not just a decision boundary. To investigate the potential benefits of incorporating uncertainty in improving probability calibration, we compared the calibrated scores from MIST algorithm with those from a vanilla isotonic regression (IST) baseline.  MIST fits a separate isotonic regression for each uncertainty level while IST  fits a single vanilla isotonic regression on the model score. 
%and performs decision boundary estimation on the regressed score similar to ST.
We evaluate the Expected Calibration Error in the $j^{th}$ score-bin, $ECE@j$ as 
\begin{equation*}
    \centering
    ECE@j = \frac{1}{K} \sum_{i \in [1,K]}  \frac{1}{n(i,j)} \bigg \lvert\sum_{\x \in Bin(i,j)} (score[\x] - label[\x]) \bigg \rvert,
\end{equation*}
where for each bin $(i,j)$,  calibration error (CE) is evaluated on samples from the bin. CE is the absolute value of the average difference between the score and  label for each sample $\x \in Bin(i,j)$, where $j \in [1,L], i \in [1,K]$. $n(i,j)$ is the number of samples in the $Bin(i,j)$ . For both MIST and IST, we use the respective isotonic score for CE calculation. In Fig. \ref{fig:ECE} (a) and (c), we plot $ECE@j$ for all score bins for MIST and IST decision boundary algorithms on Criteo and Avazu datasets respectively, averaged over 5 different experiment seeds. The difference between $ECE@j$ MIST vs IST is pronounced at high-score levels, aligning with our primary observation that leveraging  uncertainty estimates in decision boundary estimation helps improve recall at high precision levels. 

We also define the cumulative-$ECE@j$ as the averaged calibration error for all bins with model-score percentile greater than $j$. The cumulative-$ECE@j$ results in a smoothened  plot and the difference between the cumulative-$ECE@j$ for IST with that of MIST increases with model-score.

\section{Posterior Networks}
\label{appendix:posterior}
 Posterior Network (PostNet) \citep{posterior_network} builds on the idea of training a model to predict the parameters of the posterior distribution for each input sample. For classification, the posterior distribution (assuming conjugacy with exponential family distribution) would be Dirichlet distribution, and PostNet estimates the parameters of this distribution using Normalising Flows.

 % For exponential family, it is easy to see that the posterior parameters are a convex combination of the prior and the observed evidence (modeled by likelihood). The convex-weights are proportional to the pseudo-samples used in prior and likelihood as below:

They model this by dividing the network into two components:
\begin{itemize}
    \item \textbf{Encoder}: For every input $\x$, encoder ($f_{\theta}$) computes $z = f_{\theta}(\x)$, a low-dimensional latent representation of the input sample in a high-dimensional space, capturing relevant features for classification. The encoder also yields sufficient statistics of the likelihood distribution in the form of affine-transform of $z(\x)$ followed by application of log-softmax. %This forms $\alpha^i$ part of the above equation. For classification setting, the output will be the parameters of the Dirichlet Distribution. Thus 
    Instead of learning a single-point classical softmax output, it learns a posterior distribution over them, characterized by Dirichlet distribution. 
    
    \item \textbf{Normalizing flow (NF)}: This models normalized probability density $p(z|c,\phi)$ per class on the latent space $z$, intuitively acting as class conditionals in the latent space. The ground truth label counts along with normalized densities are used to compute the final pseudo counts. Thus, the component yields the likelihood evidence that is then combined with the prior to obtain the posterior for each sample. 
    
    %to parameterize known distributions by a flexible, yet tractable family: normalizing flows, since they are capable of modeling any continuous distribution given an expressive and deep enough model.
    
   % the per-class distribution of z over the entire space of z observed during training, i.e.. This component thus yields the likelihood evidence that is then combined with the prior to obtain the posterior, for each sample i. Modeling this distribution ensures that our posterior
\end{itemize}
% \begin{figure}
%     \centering
%     \includegraphics[width=9cm]{posteriormodelfig.png}
%     \caption{Overview of the posterior networks model}
%     \label{fig:posteriormodel}
% \end{figure}
% \includegraphics{posteriormodelfig}

The model is trained using an uncertainty aware formulation of cross-entropy. Here $\theta$ and $\phi$ are the parameters of the encoder and the NF respectively. Since both the encoder network $f_\theta$ and the normalizing flow parameterized by $\phi$ are fully differentiable, we can learn their parameters jointly in an end-to-end fashion. $q(\x)$ is the estimated posterior distribution over $p(\y|\x)$. The model’s final classification prediction is the expected sufficient statistic and the uncertainty is the differential entropy of the posterior distribution. The model is optimised using stochastic gradient descent using loss function that combines cross entropy with respect to true labels and the entropy of $q(\x)$.
%Eqn \ref{eq:posterior}. %\footnote{}

\begin{comment}
\begin{center}
    \begin{equation}
        \min_{\theta, \phi} \frac{1}{N}\Sigma_{i=1}^N \mathbb{E}_{q(p(y|x^i))} [CE(p(y|x^i), y^i)] - H(q^i)
       \label{eq:posterior}
    \end{equation}
\end{center}
\end{comment}
\section{Estimation Bias Analysis: Proofs of Theorems}
\label{appendix:new_analysis}
\subsection{Data Generation Process}
\label{sec:datagen}
% \todo{SM: Add the dependencies figure}
% We describe the data generation process below with
% dependencies as shown in Figure \ref{fig:dependencies}
The true positivity rate $S^{true}(\x)$ is generated from a global Beta prior with parameters $\beta_1^T$ and $\beta_0^T$, i.e., 
$$S^{true}(\x) \sim Beta(\beta_1^T, \beta_0^T).$$
The train and test samples at an input region $\x$ (modeled in a discrete fashion) are generated from the true positivity rate following a Bernoulli distribution with the negative train samples being undersampled by factor $\tau$. Let $N^{train}(\x)$ and $N^{test}(\x)$ denote the number of train and test samples at $\x$. Let $N_c^{train}(\x)$ and $N_c^{test}(\x),~ c \in \{0, 1\}$ denote the class-wise counts. The positive counts for the train and test count are given by
$$N_1^{test}(\x) \sim Binomial(N^{test}(\x),S^{true}(\x))$$
$$ N_1^{train}(\x) \sim Binomial(N^{train}(\x),\frac{\tau S^{true}(\x)}{ (\tau-1)S^{true}(\x) +1}).$$
The train and test positivity rates are given by $S^{train}(\x) =\frac{N_1^{train}(\x)}{N^{train}(\x)}$ and $S^{test}(\x) =\frac{N_1^{test}(\x)}{N^{test}(\x)}$. The model score $S^{model}(\x)$ is obtained by fitting a model on the train set with no additional dependence on the test and true positivity rates. Fig. \ref{fig:dependencies} shows the dependencies among the different variables.

\begin{lemma} 
The relationship between train positivity $S^{train}(\x)$ and model score for  positive class $S^{model}(\x)$ from Posterior Network is given by
$$ S^{train}(\x) = S^{model}(\x) -  (\omega - S^{model}(\x)) \gamma(\x).$$ 
where
 \begin{itemize}
\item  $\omega =\frac{\beta_1^{P}}{\beta_1^{P} + \beta_0^{P}}$
\item $\gamma(\x) = \frac{\beta_1^{P} + \beta_0^{P}} {\beta_1(x) + \beta_0(x)}$ 
\label{thm:model_train}
\end{itemize}
\vspace{-2mm}
\end{lemma}
\proof
Using the notation in Sec. \ref{sec:analysis}, the 
pseudo-counts $\beta_c(\x), ~c \in \{0,1\} $ correspond to 
the observed positive and negative counts at $\x$.
Hence, the train positivity is given by
$$S^{train}(\x) = \frac{\beta_1(\x) }{\beta_1(\x) + \beta_0(\x)}.$$

This gives us $\beta_0(\x) =  \beta_1(\x) \big(\frac{ 1- S^{train}(\x)} {S^{train}(\x)}\big)$.

Using the definitions of $\omega$ 
and $\gamma(\x)$, the model score $S^{model}(\x)$ from Posterior Network
(Eqn. \ref{eqn:smodel} can now be expressed in terms of $\omega, S^{train}(\x)$ and $\gamma(\x)$ as follows:
\begin{equation*}
S^{model}(\x) =  \frac{ \beta_1^{P} +  \beta_1(\x) } { \sum_{c \in \cC} [\beta_c^{P} +  \beta_c(\x)] } 
 = \frac{ \omega\gamma(\x) + S^{train}(\x)} {1+ \gamma(\x)}.
% \label{ref:eqn4}
\end{equation*}
Hence, $S^{train}(\x) = S^{model}(\x) - (\omega -S^{model}(\x))\gamma(\x)$.
\qed \\

\begin{theorem}
\label{thm:train_true_gentau}
For the case where data is generated as per Fig. \ref{fig:dependencies} and negative class is  undersampled at the rate $\frac{1}{\tau}$, the following results hold:\\
(a) The expected true positivity rate conditioned on the train positivity is given by  the expectation of the distribution,  
$$ Q(r)  = \frac{C}{(1+(\tau-1) r)^n} Beta(n(\xi\lambda(\x) + S^{train}(\x)),n( (1-\xi)\lambda(\x) + 1 - S^{train}(\x) )). $$
\begin{itemize}
\item  $n = \beta_1(\x) + \beta_0(\x)$ denotes evidence, 
~$C$ is a normalizing constant, ~$\xi =\frac{\beta_1^{T}}{\beta_1^{T} + \beta_0^{T}}$ is the positive global prior, and $\lambda(\x) = \frac{\beta_1^{T} + \beta_0^{T}} {\beta_1(\x) + \beta_0(\x)}$ is  the ratio of global priors to evidence.
\end{itemize}

(b) When there is no differential sampling, i.e., $\tau=1$, the expectation has a closed form and is given by
$$ E [ S^{true}(\x) | S^{train}(\x) ] = \frac{S^{train}(\x) + \xi\lambda(\x) }{1 + \lambda(\x)}.$$

\end{theorem} 

\proof

Let $N^{train}(\x)$and $N_1^{train}(\x)$  denote the number of train samples and positive samples associated with any input region $\x$. Then the train positivity $S^{train}(\x) = \frac{N_1^{train}(\x)}{N^{train}(\x)}$.

Since $N^{train}(\x)$ corresponds to the probability mass and pseudo counts at $\x$, we consider regions with a fixed  size $N^{train}(\x)= n$. 
The expected true positivity rate for all $\x$ with size $N^{train}(\x)= n$ conditioned on $S^{train}(\x) = \frac{k}{n}$ is given by 
$E [ S^{true}(\x)| S^{train}(\x)=k/n] = E [ S^{true}(\x)| N_1^{train}(\x)=k]$. 

For brevity, we omit the explicit mention of the dependence on $\x$ for variables  $S^{model}(\x)$, $S^{train}(\x)$, $S^{test}(\x)$, and $S^{true}(\x)$. 

The conditional probability $p(S^{true}|N_1^{train}=k)$ is given by the Bayes rule. Specifically, we have 
$$p(S^{true}=r|N_1^{train}=k) = \frac{p(S^{true}=r)p(N_1^{train}=k|S^{true}=r)}{p(N_1^{train}=k)}.$$

Here $S^{true}$ follows a global Beta prior 
and  $N_1^{train}$ a Binomial distribution with downsampling of negative examples at the rate $\frac{1}{\tau}$. For $S_{true} =r$, the success probability of the Binomial distribution (probability of obtaining a sample with $y=1$) is given by $\frac{r}{r + \frac{1-r}{\tau}}  = \frac{\tau r}{(1+ (\tau-1)r)} $. Hence, 
\begin{eqnarray*}
p(S^{true}=r)p(N_1^{train}=k|S^{true}=r) &=&
Beta(\beta_1^T,\beta_0^T)\binom{n}{k} \bigg(\frac{\tau r}{1+ (\tau-1)r} \bigg)^{k} \bigg(\frac{1-r}{1+ (\tau-1)r} \bigg)^{n-k} \\
 &=& \frac{C_0\tau^k} {{ (1+ (\tau-1)r)}^n }Beta(\beta_1^T +k ,\beta_0^T + n-k),
\end{eqnarray*}
where $C_0$ is a normalizing constant independent of $r$ and $\tau$. While  the integral $\int_r p(S^{true}=r)p(N_1^{train}=k|S^{true}=r)dr$ over $[0,1]$ does not have a closed form, we do observe that the desired conditioned distribution will have a similar form with a different normalizing constant $C$ since the denominator is independent of $r$:
\begin{eqnarray*}
p(S^{true}=r|N_1^{train}=k)&=& \frac{p(S^{true}=r)p(N_1^{train}=k|S^{true}=r)}{ \int_r p(S^{true}=r)p(N_1^{train}=k|S^{true}=r)dr}\\
&=&  \frac{C} {(1+ (\tau-1)r)^n} Beta(\beta_1^T +k ,\beta_0^T + n-k).
\end{eqnarray*}

The expected true positivity rate conditioned on $N_1^{train}=k$ is the mean of this new distribution, which does not have a closed form but 
can be numerically computed and will be similar to the simulation results in Fig. \ref{fig:biasanalysis}(b). 

Using the definitions of $\xi = \frac{\beta_1^{T}}{\beta_1^{T} + \beta_0^{T}}$ and 
$\lambda(\x) = \frac{\beta_1^{T} + \beta_0^{T}} {\beta_1(\x) + \beta_0(\x)}$,  we can rewrite  $\beta_1^T = n\xi\lambda(\x)$ and $\beta_0^T = n(1-\xi)\lambda(\x)$. Further, observing that $S^{train}(\x)=k/n$, 
we can express this distribution as 
$$Q(r)= \frac{C}{(1+(\tau-1) r)^n} Beta(n(\xi\lambda(\x) + S^{train}(\x)),n( (1-\xi)\lambda(\x) + 1 - S^{train}(\x) )), $$
which yields the desired result.\\

{\em Part b:}
For the case where $\tau =1$, the term $\frac{1}{(1+(\tau-1) r)^n}=1$ and the distribution $Q(r)$ reduces to just the Beta distribution 
$Beta(n(\xi\lambda(\x) + S^{train}(\x)),n( (1-\xi)\lambda(\x) + 1 - S^{train}(\x) ))$. Note the normalizing constant $C=1$ since the Beta distribution itself integrates to 1. The expected true positivity is the just the mean of this Beta distribution, i.e.,
\begin{eqnarray*}
 E [ S^{true}(\x) | S^{train}(\x) ] &=& 
 \frac{n(\xi\lambda(\x) + S^{train}(\x))}{n(\xi\lambda(\x) + S^{train}(\x))  +n( (1-\xi)\lambda(\x) + 1 - S^{train}(\x) )}\\
 &=& \frac{S^{train}(\x) + \xi\lambda(\x) }{1 + \lambda(\x)}.
 \end{eqnarray*}
 
\qed \\

\begin{theorem} 
For the case where data is generated as per Sec. \ref{sec:datagen}, 
the expected test and true positivity rate conditioned either on the train positivity rate or model score for positive class are equal, i.e.,
$$E[S^{test}(\x)|S^{train}(\x)]  =  E[S^{true}(\x) | S^{train}(\x)], $$
$$E[S^{test}(\x)|S^{model}(\x)]  =  E[S^{true}(\x) | S^{model}(\x)]. $$
\label{thm:true_test}
\vspace{-2mm}
\end{theorem}
\proof
From the data generation process in Sec. \ref{sec:datagen}, we observe that the test label samples $Y_{test}(\x)$ at the input region $\x$ are generated by Bernoulli distribution centered around $S^{true}(\x)$ 
i.e., $Y_{test}(\x) \sim Bernoulli(S^{true}(\x))$ and $S^{test}$ is the mean of $Y^{test}$ over the test samples.   

Hence, the test labels $Y^{test}(\x)$ and test positivity rate  $S^{test}(\x)$  are  independent of the model score $S^{model}(\x)$ and $S^{train}(\x)$ given the $S^{true}(\x)$.  For brevity, we omit the explicit mention of the dependence on $\x$  for variables $Y^{test}(\x)$, $S^{model}(\x)$, $S^{train}(\x)$, $S^{test}(\x)$, and $S^{true}(\x)$.

As $Y^{test}$ is  conditionally independent of  $S^{train}$ given $S^{true}$, we observe that
$$E[Y^{test}|S^{train}] = E_ {S^{true} |S^{train}} [  E [Y^{test}|S^{true} ]  ]. $$

However, since  $Y_{test} \sim Bernoulli(S^{true}(\x))$, we have $E [Y^{test}|S^{true} ]   = S^{true}$.  Therefore, 
$$E[Y^{test}|S^{train}] = E_ {S^{true} |S^{train}} [  E [Y^{test}|S^{true} ]   ]. =E[S^{true} | S^{train}]$$

Since $S^{test}$ is itself the expectation  over $Y_{test}$,  by the law of iterated expectations, we have,

$$E[S^{test}|S^{train}]  = E[Y^{test}|S^{train}] =  E[S^{true} | S^{train}], $$
which is the desired result.  The same result holds true even when conditioning on the model score $S^{model}$ since $Y^{test}$ and $S^{test}$ are also  conditionally independent of  $S^{model}$ given $S^{true}$.

\qed \\

\begin{theorem}  {\bf [Restatement of Theorem \ref{thm:bias}]}
\label{thm:bias2}
For the case where data is generated as per Fig. \ref{fig:dependencies} and negative class is  undersampled at the rate $\frac{1}{\tau}$:\\
%the following results hold:\\
(a) The expected test and true positivity rate conditioned on the train positivity  are equal and correspond to the expectation of the distribution,  
$$ Q(r)  = \frac{C}{(1+(\tau-1) r)^n} Beta(n(\xi\lambda(\x) + S^{train}(\x)),n( (1-\xi)\lambda(\x) + 1 - S^{train}(\x) )). $$

When there is no differential sampling, i.e., $\tau=1$, the expectation has a closed form and is given by
$$ E [ S^{true}(\x) | S^{train}(\x) ] = E [ S^{test}(\x) | S^{train}(\x)] = \frac{S^{train}(\x) + \xi\lambda(\x) }{1 + \lambda(\x)}.$$
\begin{itemize}
\item  $n = \beta_1(\x) + \beta_0(\x)$ denotes evidence, 
~$C$ is a normalizing constant, ~$\xi =\frac{\beta_1^{T}}{\beta_1^{T} + \beta_0^{T}}$ is the positive global prior, and $\lambda(\x) = \frac{\beta_1^{T} + \beta_0^{T}} {\beta_1(\x) + \beta_0(\x)}$ is  the ratio of global priors to evidence.
%\item  $\omega =\frac{\beta_1^{P}}{\beta_1^{P} + \beta_0^{P}}$ and $\xi =\frac{\beta_1^{T}}{\beta_1^{T} + \beta_0^{T}}$  are the positive prior ratios,  
%\item $\nu = \frac{\beta_1^{T} + \beta_0^{T}} {\beta_1^P + \beta_0^P}$ is the ratio of global and model priors.
%_{\text{true-priors : model-priors}}$ 
%\item $\nu = \underbrace{ \frac{\beta_1^{T} + \beta_0^{T}} {\beta_1^P + \beta_0^P}}$ is the ratio of global and model priors.
%_{\text{true-priors : model-priors}}$ 
\\
\end{itemize}
(b) For Posterior Networks,  the test and true positivity rate conditioned on the model score $S^{model}(\x)$ can be obtained using $S^{train}(\x) = S^{model}(\x) - (\omega - S^{model}(\x)) \gamma(\x)$. For $\tau =1$, the estimation bias, i.e., difference between model score and test positivity is given by $\frac{(S^{model}(\x)(\nu-1) +\omega -\xi\nu)\gamma(\x)}{1+\nu\gamma(\x)}.$
%, where
\begin{itemize}
    \item $\omega =\frac{\beta_1^{P}}{\beta_1^{P} + \beta_0^{P}}$ %is the positive model prior, 
    and  $\nu = \frac{\lambda(\x)}{\gamma(\x)} =\frac{\beta_1^{T} + \beta_0^{T}} {\beta_1^P + \beta_0^P}$ is the ratio of global and model priors.
%_{\text{true-priors : model-priors}}$ 
\end{itemize}
\end{theorem}
\proof

{\em Part a:}
From Theorem \ref{thm:train_true_gentau}, we directly obtain the result on the expectation of true positivity rate in terms of the train positivity both for the general case where $\tau \neq 1$ and for the special case of $\tau =1$.  Further, from Theorem \ref{thm:true_test}, we observe that the expected true positivity is also the same as the expected test positivity conditioned on the train positivity, which yields the desired result.\\

{\em Part b:}\\
From Lemma \ref{thm:model_train}, we obtain the relationship between the train positivity and the model score, i.e., 
$S^{train}(\x) = S^{model}(\x) - (\omega - S^{model}(\x)) \gamma(\x)$.
which can be used to expression the expected train and test positivity directly in terms of the model score.

For the case $\tau=1$, in particular, since $S^{train}(\x)$ is deterministic function of $S^{model}(\x)$ for a fixed $\gamma(\x)$, we observe that
$$E [ S^{test}(\x) | S^{train}(\x)] = \frac{S^{train}(\x) + \xi\lambda(\x) }{1 + \lambda(\x)}.$$
Expressing this in terms of $S^{model}(\x)$ and $\gamma(\x) = \lambda(\x)/\nu$ gives us
 $$ E [ S^{test}(\x) | S^{model}(\x)] =
 \frac{S^{model}(\x) + ( S^{model}(\x) + \xi\nu -\omega)\gamma(\x)}{1+\nu\gamma(\x)}. $$ 
Thus, the estimation bias is given by 

$$S^{model}(\x) - E [ S^{test}(\x) | S^{model}(\x)] = \frac{(S^{model}(\x)(\nu-1) +\omega -\xi\nu)\gamma(\x)}{1+\nu\gamma(\x)}.$$
\qed \\

\section{Computational Complexity of Decision Boundary  Algorithms}
\label{appendix:nphard}

\begin{lemma}
Given a $K \times L$ grid with positive sample counts $[p(i,j) ]_{K \times L}$ and total sample counts $[n(i,j) ]_{K \times L}$ and any boundary $\myb=[b(i)]_{i=1}^K$ that satisfies  $precision(\myb) \geq \sigma$ and $recall(\myb) \geq \eta$, let $b^{chp}(i)$ denote the minimum score threshold $j$ such that $\frac{p(i,j')}{n(i,j')} \geq \sigma$ for all $j' \geq j$, i.e., contiguous high precision region.  Then, then the new boundary $\myb'$ defined as  $b'(i) = \text{min} ~(b(i), b^{chp}(i) ), ~~ \forall [i]_1^K$ also satisfies $precision(\myb) \geq \sigma$ and $recall(\myb) \geq \eta$.
\label{lem:pos_bins}
\end{lemma}
\proof

Let $B'^{+}$ denote the positive region for the new boundary $\myb'$ and $B^{chp}$ the contiguous high precision bins for each uncertainty level, i.e., 
$B^{chp} = \{(i,j) | j >b^{chp}(i), \forall 
[i]_1^K, ~ [j]_0^L\}$.

By definition, we have,
%\begin{equation*}
$B'^{+} = \{(i,j) | j >b'(i), ~ \forall [i]_1^K, ~ [j]_0^L \} =  B^{+} \bigcup B^{chp}.$
%\end{equation*}
Given a set of bins $B$, let $P(B)$ and $N(B)$ denote the net positive and total samples within this set of bins. Since $precision(\myb) \geq \sigma$, we have $P(B^{+}) \geq \sigma N(B^{+})$.  Since $\frac{p(i,j)}{n(i,j)} \geq \sigma, ~\forall (i,j) \in B^{chp}$, we also note that  $P(B) \geq \sigma N(B)$ for any set $B \subseteq B^{chp}$. 

Now, the precision for the new boundary is given by
\begin{eqnarray*}
precision(\myb') & = &  \frac{ \sum_{(i,j) \in  B'^{+}} p(i,j)} {\sum_{(i,j) \in  B'^{+}}  n(i,j) } 
= \frac{ \sum_{(i,j) \in  B^{+}} p(i,j) + \sum_{(i,j) \in  B^{chp} \setminus B^{+}} p(i,j)} 
{ \sum_{(i,j) \in  B^{+}} n(i,j) + \sum_{(i,j) \in  B^{chp} \setminus B^{+}} n(i,j)}  \\
&=& \frac{P(B^{+}) + P(B^{chp} \setminus B^{+})}{N(B^{+}) + N(B^{chp} \setminus B^{+})} \\
& <& \sigma \bigg ( \frac{N(B^{+}) + N(B^{chp} \setminus B^{+})}{N(B^{+}) + N(B^{chp} \setminus B^{+})} \bigg) ~~ \{ \text{since} ~~ (B^{chp} \setminus B^{+}) \subseteq B^{chp}  \} \\
 &=& \sigma
\end{eqnarray*}
Let $P_0$ denote the total number of positive samples. Then the recall for the new boundary is given by 
$$recall(\myb') =  \frac{ \sum_{(i,j) \in  B'^{+}} p(i,j)} { P_0}  = \frac{P(B^{+}) + P(B^{chp} \setminus B^{+})}{P_0} \geq  \frac{P(B^{+})}{P_0} > \eta.$$
Hence, $\myb'$ also satisfies the precision and recall bounds.\qed\\

\begin{theorem}{\bf [Restatement of Theorem \ref{thm:nphard}]}
The problem of computing the optimal 2D- binned decision boundary (2D-BDB)  is NP-hard. 
\end{theorem}

\proof  The result is obtained by demonstrating that any instance of  the well-known subset-sum problem defined below can be mapped to a specific instance of a reformulated 2D-BDB problem such that there exists a solution for the subset-sum problem instance if and only if there exists a solution for the equivalent decision boundary problem.  

Specifically, we consider the following two problems:

\underline{\it Subset-sum problem}: Given a finite set
$\cA = \{a_1, \ldots, a_t \}$ of $t$ non-negative integers and a target sum $T$, is there a subset $\cA^{’}$ of $\cA$ such that
$\sum_{a_r\in \cA^{’}} a_r = T$. \\

\underline{\it Reformulated 2D-BDB problem}: Given a $K\times L$ grid with $p(i,j)$ and $n(i,j)$ denoting the positive and total number of samples for bin $(i,j)$, is there a decision boundary 
$ \myb = [b(i)]_{i=1}^K$ such that $precision(\myb) \geq \sigma$ and $recall(\myb) \geq \eta$. \\

Let $B^{+}$ denote the positive region of the boundary, i.e., $B^{+} = \{ (i,j) | 1\leq i \leq K, 1 \leq j \leq L; ~ j > b(i) \}$ and $P_0$ denote the total number of positive samples. Then, we require
\begin{itemize}
\item $precision(\myb) =  \frac {\sum_{(i,j) \in B^{+}} p(i,j) } {\sum_{(i,j) \in B^{+}} n(i,j) } \geq \sigma $ 
\item $recall(\myb) =  \frac {\sum_{(i,j) \in B^{+}} p(i,j) }  {P_0} \geq \eta, $
\end{itemize}
Note that maximizing recall for a precision bound is equivalent to reformulation in terms of the existence of a solution that satisfies the specified precision bound and an arbitrary recall bound. 

Given any instance of subset-sum problem with $t$ items , we construct the equivalent decision boundary problem by mapping it to  a $(t+1) \times 1 $ grid  (i.e., $K= t+1, L=1$ with bins set up as follows.
\begin{itemize}
\item $n(i,1) = T ;~~ p(i,1) = 2\sigma T,$
\item $n(i+1, 1) = a_i; ~~ p(i+1,1) =2\epsilon a_i,$
\end{itemize}
where the  parameters $\sigma, \epsilon, \eta$  can be chosen to be any set of values that satisfy 
\begin{itemize} 
\item $0 \leq \sigma \leq \frac{1}{2},$
%\item 
$~~0 < \epsilon <  \frac{\sigma}{2(T+1)}, $
%\item  
$~~\eta  = \frac{2 (\sigma + \epsilon) T}{P_0}$.
\end{itemize}

We prove that the problems are equivalent in the sense that the solution for one can be constructed from that of the other.

{\bf Part 1: Solution to subset sum $\Rightarrow$ Solution to decision boundary }

Suppose there is a subset $\cA’$  such that $ \sum_{a_i\in \cA’} a_i = T$. Then, consider the boundary $\myb$ defined as $b(1)=1$ and $b(i)=\1[a_i \notin \cA']$, i.e., the positive $B^{+}  =  \{(1,1)\} \bigcup  \{ ( i+1, 1)  | a_i \in \cA’ \}$.
This leads to the following  precision and recall estimates.

\begin{equation*}
precision (\myb) = \frac {\sum_{(i,j) \in B^{+}} p(i,j) } {\sum_{(i,j) \in B^{+}} n(i,j) } 
=  \frac{ 2\sigma T + 2\epsilon \sum_{a_i \in \cA’} a_i}{ T + \sum_{a_i \in \cA’} a_i} \\
= \frac{2(\sigma +\epsilon)T}{2T}  \geq \sigma
\end{equation*}

\begin{equation*}
recall(\myb) = \frac {\sum_{(i,j) \in B^{+}} p(i,j) }  {P_0}  
= \frac{ 2\sigma T + 2\epsilon \sum_{a_i \in \cA’} a_i}{ P_0} 
= \frac{2(\sigma +\epsilon)T}{P_0}  =\eta.
\end{equation*}

Since this choice of $\myb$ is a valid boundary satisfying the precision and recall requirements, we have a solution for the decision boundary problem.

{\bf Part 2: Solution to decision boundary $\Rightarrow$ Solution to subset sum }

Let us assume we have a solution for the decision boundary, i.e.,  we have a boundary $\myb$ with 
$precision(\myb) \geq \sigma$ and $recall(\myb)  \geq \eta$  respectively. Since the positivity rate of the bin $(1,1)$ is $\frac{2\sigma T}{T} = 2\sigma  > \sigma$, from Lemma \ref{lem:pos_bins} we observe that the boundary $\myb$  is such that $(1,1)$ is in the positive region of the boundary $B^{+}$.

Consider the subset $\cA’ = \{a_i |  (i+1,1) \in B^{+} \}$.  We will now prove that $ \sum_{a_i \in \cA’} = T$ which makes it a valid solution for the subset-sum problem. 

Suppose that $ \sum_{a_i \in \cA’} = T’ >T$, i.e., $T’ \geq T+1$ since T is an integer. For this case, we have 
\begin{equation*}
 precision (\myb) = \frac {\sum_{(i,j) \in B^{+}} p(i,j) } {\sum_{(i,j) \in B^{+}} n(i,j) }  \\ 
=  \frac{ 2\sigma T + 2\epsilon \sum_{a_i \in \cA’} a_i}{ T + \sum_{a_i \in \cA’} a_i} \\
= \frac{2\sigma T +2\epsilon T’}{T+T’}. 
\end{equation*}

Since $\epsilon <  \frac{\sigma}{2(T+1)}$, we have 
\begin{eqnarray*}
precision(\myb) &= &\frac{2\sigma T +2\epsilon T’}{T+T’}
  <  \frac{2 \sigma T + \frac{2\sigma T’}{2(T+1)}}{T+T’} 
 =  \sigma \bigg(\frac{2T  + \frac{T’}{T+1}}{T+T’} \bigg)
 =  \sigma \bigg (1 +  \frac{T-T’  + \frac{T’}{T+1}}{T+T’} \bigg) \\
 &=&  \sigma \bigg (1 +  \frac{T  - \frac{TT’} {T+1}}{T+T’} \bigg)
 =  \sigma \bigg (1 -  \frac{T(T’ - T-1)} {(T+1)(T+T’)} \bigg)
 \leq  \sigma.   ~~ \{ \text{since}~~ T’ \geq T+1 \} \\
\end{eqnarray*} 
In other words, $ precision (\myb) <\sigma $, which is a contradiction since $\myb$ is a valid solution to the decision boundary problem.

Next consider the case where  $ \sum_{a_i \in \cA’} = T’ < T$.  Then, we have, 

\begin{equation*}
recall(\myb) = \frac {\sum_{(i,j) \in B^{+}} p(i,j) }  {P_0} 
=  \frac{ 2\sigma T + 2\epsilon \sum_{a_i \in \cA’} a_i}{ P_0} 
= \frac{2\sigma T+2\epsilon T’}{P_0}  
<  \frac{2\sigma T+2\epsilon T}{P_0} 
=  \eta.
\end{equation*}

This again leads to a contradiction since $\myb$ is a solution to the  decision boundary problem requiring $recall(\myb) \geq \eta$. Hence, the only possibility is that $ \sum_{a_i \in \cA’} = T$, i.e., we have a solution for the subset-sum problem. Since the subset-sum problem is NP-hard~\citep{subsetsum}, from the reduction, it follows that the 2D-BDB problem is also NP-hard.  \qed \\

\section{Decision Boundary Algorithms}
\label{appendix:algorithm}

Here, we provide additional details on the following proposed algorithms from Sec. \ref{sec:algo} that are used in our evaluation. These are applicable  for both variable or equi-weight binning scenarios.

\noindent {\bf Equi Weight DP-based Multi-Threshold algorithm (\textsc{EW-DPMT})} : We detail the EW-DPMT (Algorithm \ref{alg:algo3}), presented in Sec. \ref{sec:algo} here.
Let  $R(i,m), ~[i]_1^K, [m]_0^{KL}$  denote the maximum true positives for any decision boundary over the sub-grid with uncertainty levels $1$ to $i$ and entire score range, such that the boundary has exactly $m$ bins in its positive region.
Further, let $b(i,m,:)$  denote the optimal boundary that achieves this maximum with $b(i,m,i')$ denoting the boundary position for the $i' (\leq i)$ uncertainty level.  
For the base case when $i=1$, there is a feasible solution only for $0 \leq m \leq L$ which is the one  corresponding to $b(1,m,1) = L - m$, since the  score threshold index for picking $m$ bins in the positive region will be $L-m$. 
Now, for the case $i>1$,  we can decompose the estimation of maximum recall as follows. 
Let $j$ be the number of bins chosen as part of positive region from the $i^{th}$ uncertainty level, 
then the budget available for the lower $(i-1)$ uncertainty levels is exactly $m-j$. Hence, we have,
% \begin{equation}
$R(i, m) = \underset{0 \leq j \leq L}{\text{max}} [\pi(i,j)  +  R(i-1, m-j)]$, 
% \end{equation}
where $\pi(i,j)= \sum_{j'=L-j+1}^{L} p(i,j')$, i.e., the sum of the positive points in the $j$ highest score bins. The optimal boundary $b(i,m,:)$ is obtained by setting  $b(i,m,i) = L- j^*$ and the remaining thresholds to that of $b(i-1, m-j^*, :)$  where $j^*$ is the optimal choice of $j$ in the above recursion. 

Performing this computation progressively for all uncertainty levels and positive region bin budgets yields maximum recall over the entire grid for each choice of bin budget. This is equivalent to obtaining the entire PR curve and permits us to pick the optimal solution for a given precision bound. 
Since the bin-budget can go up to $KL$ and the number of uncertainty levels is $K$, the number of times the maximum recall optimization is invoked is $K^2L$.  The optimization itself explores $L$ choices, each being a $O(1)$ computation since the cumulative sums of positive bins can be computed progressively. Hence, the overall algorithm has $O(K^2L^2)$ time complexity and $K^2L$ storage complexity. Algorithm \ref{alg:algo3}\textsc{Equi-Weight DP-based Multi-Thresholds (EW-DPMT)} shows steps for computing the optimal 2D-decision boundary. Note that if a solution is required for a specific precision bound $\sigma$, then complexity can be reduced by including all contiguous high score bins with positivity rate $\geq \sigma$ since those will definitely be part of the solution (Lemma \ref{lem:pos_bins}).

\noindent {\bf Variable Weight DP-based Multi-Threshold algorithm (\textsc{VW-DPMT})}  As discussed in Sec. \ref{sec:algo}, the general case of the 2D-BDB problem with variable-sized bins is NP-hard, but it permits a pseudo-polynomial solution using a dynamic programming approach. Similar to the equi-weight DP algorithm \textsc{EW-DPMT}, we track the maximum recall solutions of sub-grids up to $i^{th}$ uncertainty level with a budget over the number of positive samples.  
  
Let  $R^{var}(i,m)$  denote the maximum true positives for any decision boundary over the sub-grid with uncertainty levels $1$ to $i$ and the entire score range such that the boundary has exactly $m$ samples in its positive region. We can then use the decomposition, 
\begin{equation*}
\vspace{-1mm}
R^{var}(i, m) = \underset{0 \leq j \leq L}{\text{max}} [ \pi(j) +  R^{var}(i-1, m-\nu(j) )],
\label{eqn:r_var}
\vspace{-13mm}
\end{equation*}
where $\pi(i,j)= \sum_{j'=L-j+1}^{L} p(i,j')$ and $\nu(i,j)= \sum_{j'=L-j+1}^{L} n(i,j')$. 

Algorithm \ref{alg:algo5} provides details of the implementation assuming a dense representation for the matrix $R^{var}$ (Eqn. \ref{eqn:r_var}) that tracks all the maximum true positive (i.e., unnormalized recall) solutions for sub-grids up to different uncertainty levels and with a budget on the number of samples assigned to the positive region. For our experiments, we implemented the algorithm using a sparse representation for $R^{var}$ that only tracks the feasible solutions.

\noindent {\bf  Greedy Multi-Thresholds (\textsc{GMT}) } Algorithm \ref{alg:algo2} provides the details of this greedy approach 
where we independently choose the score threshold for each uncertainty level. Since all the score bin thresholds are progressively evaluated for each uncertainty level, the computational time complexity is $O(KL)$ and the storage complexity is just $O(K)$.
However, this approach can even be inferior to the traditional approach of picking a single global threshold on the score, which is the case corresponding to a single uncertainty level. \textsc{ST} algorithm can be viewed as a special case of \textsc{GMT} algorithm where only one uncertainty level is considered (i.e. $K=1$).

\noindent {\bf Multi Isotonic regression Single Threshold (\textsc{MIST}) } As mentioned earlier, the isotonic regression-based approach involves performing isotonic regression~\citep{isotonic} on each uncertainty level to get calibrated scores that are monotonic with respect to the score bin index. Bins across the entire grid are then sorted based on the calibrated scores and a global threshold on the calibrated score that maximizes recall while satisfying the desired precision bound is picked. In our implementation, we use the isotonic regression implementation is \texttt{scikit-learn}, which has linear time in terms of the input size for  $L_2$ loss~\citep{isotonicstout}.  Since the sorting based on calibrated scores is the most time consuming part, this algorithm has a time complexity of $O(KL\log(KL))$ and a storage complexity of $O(KL)$. For our experiments, we performed isotonic regression for each of the $K$ uncertainty levels directly using the samples instead of the aggregates at $L$ score bins. When the $K$ uncertainty bins are equi-weight, this is essentially the case where $L=N/K$.

\begin{algorithm}[H]
\caption{Greedy Decision Boundary - Multiple Score Thresholds [\textsc{GMT}]}
 \label{alg:algo2}
 \begin{algorithmic}
  \STATE \hspace*{-4mm} {\bf Input:} Variable-sized $K \times L$ grid with positive sample counts $[p(i,j) ]_{K \times L}$ and total sample counts   $[n(i,j) ]_{K \times L}$, overall sample count $N$, precision bound $\sigma$.
  \STATE \hspace*{-5mm} {\bf Output:} (unnormalized) recall $R^*$ and corresponding boundary $\myb^*$ for precision $\geq \sigma$ with greedy approach.
  \STATE \hspace*{-5mm} {\bf Method:}
  \STATE  \textit{// Pre-computation of cumulative sums of positives}
  \FOR{$i=1$ to $K$}
  \STATE $\pi(i,0) = 0$
  \STATE $\nu(i,0) = 0$
  \FOR{$j=1$ to $L$}
  \STATE $\pi(i,j) = \pi(i,j-1) + p(i, L-j+1)$
  \STATE $\nu(i,j) = \nu(i,j-1) + n(i, L-j+1)$
  \ENDFOR
  \ENDFOR
  \STATE \textit{// Initialization}
  \STATE $R =  0$  
  \STATE \textit{// Independent Greedy Score Thresholds}
  \FOR{$i=1$ to $K$}
  \STATE $j^* =  \underset{0 \leq j \leq L, ~  s.t. \frac{\pi(i,j)}{ \nu(i,j)}  \geq \sigma } {\text{argmax}}  [ \pi (i,j) ] $
  \STATE $b(i) =j^*$
  \STATE $R = R + \pi(i,j^*)$
  \ENDFOR
 \STATE $R^* = R , ~ \myb^* = b(:) $
 \RETURN $(R^*, \myb^*)$ %
%  \STATE \textbf{return} $(R^*, \myb^*) \quad$ %
%  \STATE \RETURN $(R^*, \myb^*)$
%  \STATE {\bf return} $(R^*, \myb^*)$
 \end{algorithmic}
 \end{algorithm}

\begin{algorithm}[H]
\caption{Greedy Decision Boundary - Global Threshold on Score Recalibrated with Isotonic Regression [\textsc{MIST}]}
 \label{alg:algo7}
 \begin{algorithmic}
  \STATE \hspace*{-4mm} {\bf Input:} Variable-sized $K \times L$ grid with positive sample counts $[p(i,j) ]_{K \times L}$ and total sample counts 
  $[n(i,j) ]_{K \times L}$, overall sample count $N$,
  precision bound $\sigma$.
  \STATE \hspace*{-5mm} {\bf Output:} (unnormalized) recall $R^*$ and corresponding boundary $\myb^*$ for precision $\geq \sigma$ with greedy approach.
  \STATE \hspace*{-5mm} {\bf Method:}
  \STATE  \textit{// Recalibrate each row using isotonic regression}
  \FOR{$i=1$ to $K$}
%   \STATE $X_{iso} = \bigcup_{(i,j) \forall j \in [1,L]} x, x \in Bin(i,j)$ 
%   \STATE  $[s^{iso} (x)] =$ \textsc{IsotonicRegression}$([ (y(x), score(x)]) \forall x \in X_{iso}$
  \STATE  $[s^{iso}(i,j)]_{j=1}^L =$ \textsc{IsotonicRegression}$([ (p(i,j), n(i,j))]_{j=1}^{L} )$
  \ENDFOR
  \STATE  \textit{// Get a global threshold on calibrated score}
  \STATE  \textit{// rank is descending order 0 to maxrank - low rank means high positivity }
  \STATE $[rank(i,j)]_{K \times L} =  \textsc{Sort}([s^{iso}(i,j)]_{K \times L} ) $
  \STATE $maxrank = \underset { [i]_1^K  ~ [j]_1^L}{\text{max}} rank(i,j)$
  \STATE $\pi(0) = 0, \nu(0) = 0$
  \STATE $r=0$
  \REPEAT 
  \STATE $r =r +1$
  \STATE $\pi(r) = \pi(r-1) + \sum_{ (i,j) | ~rank(i,j) =r }[s^{iso}(i,j)n(i,j)]$
  \STATE $\nu(r) = \nu(r-1) + \sum_{ (i,j) | ~rank(i,j) =r } [n(i,j)]$
  \UNTIL$\bigg( (\frac{\pi(r)} {\nu(r)} < \sigma) \vee (r > maxrank ) \bigg)$
  \STATE $r^*=r-1$
  \STATE  \textit{// Obtain score thresholds for different uncertainty levels}
  \STATE $R =  0$  
  \FOR{$i=1$ to $K$}
  \IF{$\{ j | rank(i,j) \geq r^* \} = \emptyset$}
  \STATE $j^* = L$
  \ELSE 
  \STATE $j^* = \underset{ j | rank(i,j) \geq r^* } {\text{argmin}} [j]  $
  \ENDIF
  \STATE $b(i) =j^*$
  \FOR{$j=j^*+1$ to $L$}
  \STATE $R = R + p(i, j) $
  \ENDFOR
  \ENDFOR
 \STATE $R^* = R , ~ \myb^* = b(:) $
 \STATE {\bf return} $(R^*, \myb^*)$
 \end{algorithmic}
 \end{algorithm}

\begin{algorithm}[H]
\caption{Optimal Decision Boundary for Variable-Weight Bins [\textsc{VW-DPMT}]}
 \label{alg:algo5}
 \begin{algorithmic}
  \STATE \hspace*{-4mm} {\bf Input:} Variable-sized $K \times L$ grid with positive sample counts $[p(i,j) ]_{K \times L}$ and total sample counts 
  $[n(i,j) ]_{K \times L}$, overall sample count $N$,
  precision bound $\sigma$.
  \STATE \hspace*{-5mm} {\bf Output:} maximum (unnormalized) recall $R^*$ and corresponding optimal boundary $\myb^*$ for precision $\geq \sigma$.
  \STATE \hspace*{-5mm} {\bf Method:}
  \STATE \textit{// Initialization}
  \STATE $R(i,m) =  -\infty$; $~b(i,m,i’) =  -1$;  ~$ [i]_1^K,~[i’]_1^K, ~ [m]_0^N)$
  \STATE  \textit{// Pre-computation of cumulative sums of positives}
  \FOR{$i=1$ to $K$}
  \STATE $\pi(i,0) = 0$
  \STATE $\nu(i,0) = 0$
  \FOR{$j=1$ to $L$}
  \STATE $\pi(i,j) = \pi(i,j-1) + p(i, L-j+1)$
  \STATE $\nu(i,j) = \nu(i,j-1) + n(i, L-j+1)$
  \ENDFOR
  \ENDFOR
  \STATE \textit{// Base Case: First Uncertainty Level}
  \FOR{$j=0$ to $L$}
  \STATE $m = \nu(1,j)$
  \STATE $R(1,m) = \pi(1,j)$
  \STATE $b(1,m,1) = L-j$
  \ENDFOR
 \STATE \textit{// Decomposition: Higher Uncertainty Levels}
 \FOR{$i=2$ to $K$}
  \FOR{$m=0$ to $\sum_{i'=0}^i N^{cum(i,j)}$}
  \STATE $j^* = \underset{0 \leq j \leq L}{\text{argmax}} [ \pi(i,j) +  R(i-1, m-\nu(i,j)) ]$
  \STATE $R(i, m) = \pi(i,j^*) + R(i-1, m - \nu(i,j^*))$
  \STATE $b(i,m,:)=b(i-1,m-\nu(i,j^*),:)$  
  \STATE $b(i,m,i) = L- j^* $
\ENDFOR
\ENDFOR
\STATE \textit{// Maximum Recall for Precision}
\STATE $m^* =  \underset{0 \leq m \leq KL~ s.t. \frac{R(K,m)}{m} \geq \sigma } {\text{argmax}}  [ R(K,m)] $
\STATE $R^* = R(K,m^*)$; $\myb^* = b(K,m^*,:)$
\STATE {\bf return} $(R^*, \myb^*)$
\end{algorithmic}
\end{algorithm}

\newpage
\section{Notations}
\label{appendix:notations}
\begin{table}[h]
    \centering
    \begin{tabular}{|l|l|}
    \hline
    \textbf{Symbol} & \textbf{Definition} \\
    \hline
    $\x$ & an input instance or region \\
    $y$ & target label for an input sample $\x$ \\
    $\cC$ & set of class labels $\{0,1\}$ \\
    $c$ & index over the labels in $\cC$  \\
    $[i]_{lb}^{ub}$ & index iterating over integers in $\{lb, \cdots, ub\}$ \\
    \hline
    \hline
    \multicolumn{2}{|c|}{\textit{Estimation Bias and Posterior Network Related}} \\
    \hline
    \hline
    $\pr(\cdot)$ & Probability distribution \\
    $q(\x)$ & Distribution over class posterior at $\x$ output by Posterior Network \\
     $H(q(\x))$ & differential entropy of distribution $q(\x)$ \\
    $\alpha_c(\x)$ & Parameters of Beta distribution $q(\x)$ for class $\cC$ \\
    $\beta^P_c$ & Parameters of Model prior  for class $\cC$\\
    $\beta^T_c$ & Parameters of  True prior  for class $\cC$\\
    $\beta_c(\x)$ & Pseudo counts for class $\cC$\\
    $N_c$ & observed counts for class $\cC$\\
    $\z(\x)$ & penultimate layer representation from the model \\
    $\phi$ & parameters of normalizing flow in Posterior Networks \\
    
    $u(\x)$ & Uncertainty for $\x$ \\
    $ S^{model}(\x),~ s(\x)$ & Model score for positive class \\
    $S^{true}(\x)$ & true positivity in input region $\x$ \\
    $S^{train}(\x)$ & empirical positivity in the train set for input region $\x$ \\
    $S^{test}(\x)$ & empirical positivity in the test set for input region $\x$ \\
    $\tau$ & differential sampling rate for negaatives \\
    $n=n(\x)$ & evidence at input region $\x$ given by $\beta_1(\x)+\beta_2(\x)$ \\
    $\xi$ & positive class fraction in global prior \\
    $\omega$ & positive class fraction in model prior \\
    $\lambda(\x)$ & ratio of global priors to evidence \\
    $\gamma(\x)$ & ratio of model priors to evidence \\
    $\nu$ & ratio of global priors to model priors\\
    $Q(\cdot)$ & distribution of true positivity conditioned on a fixed train positivity rate \\
    \hline
    \hline
    \multicolumn{2}{|c|}{\textit{Decision Boundary Related}} \\
    \hline
    \hline
    $D^{train}$, $D^{hold}$, $D^{test}$ & Data split for training the model, calibrating the decision boundary and testing \\
    
    $\myb$ & decision boundary defined in terms of score and uncertainty thresholds \\
    $\psi_{\myb}(\x)$ & labeling where samples that satisfy boundary thresholds are  positives \\
    $\myb(u)$ & decision boundary specified by a score threshold for a fixed uncertainty level \\
    $\cS$ & Range of score values \\
    $\cU$ & Range of uncertainty values \\
    $K$ & Number of uncertainty bins \\
    $L$ & Number of score bins \\
    $\rho(s,u)=(\rho^{\cS}(s),\rho^{\cU}(u))$ & Partitioning function that maps score and uncertainty values to bin-index $(i,j)$ \\
    % $\rho^{\cS}(s(\x))$ & score function \\
    % $\rho^{\cU}(u(\x))$ & uncertainty function \\
    $R(i,m)$ & max. true positives for any boundary upto the $i^{th}$ uncertainty level with \\
    & exactly $m$ positive bins in \textsc{EW-DPMT} \\
    $b(i,m,:)$ & max. recall  boundary for the sub-grid upto uncertainty level $i$, with exactly $m$ positive bins \\
    $p(i,j)$ & count of positives in the $(i,j)$th bin \\
    $n(i,j)$ & count of samples in the $(i,j)$th bin \\
    $\pi(i,j)$ & count of positive samples in the $j$ highest score bins for uncertainty level $i$ \\
    \hline
    \end{tabular}
    \caption{Notation and their definitions.}
    \label{tab:notation}
\end{table}

\end{document}